\documentclass[sigconf]{acmart}

\AtBeginDocument{%
 \providecommand\BibTeX{{%
    \normalfont B\kern-0.5em{\scshape i\kern-0.25em b}\kern-0.8em\TeX}}}

\usepackage{soul}
\usepackage{color}
\usepackage{wallpaper}
\usepackage{graphicx}
\usepackage{cleveref}
\usepackage{numprint}
\usepackage{paralist}
\usepackage{xspace}

\definecolor{orangeX}{rgb}{1,.5,0}
\definecolor{blueX}{rgb}{.2, .59, .88}
\definecolor{purpleX}{rgb}{.294118, 0, .509804}
\definecolor{greenX}{rgb}{.421, .578, .241}
\definecolor{bole}{rgb}{0.47, 0.27, 0.23}
\definecolor{mypink3}{cmyk}{0, 0.7808, 0.4429, 0.1412}
\definecolor{mygray}{gray}{0.6}
	
\newcommand{\musesec}{\textsc{MuSe-CaR}}
\newcommand{\ulmlong}{Ulm-Trier Social Stress\,}
\newcommand{\ulm}{\textsc{Ulm-TSST}}

\newcommand{\musewild}{\textsc{MuSe-Wilder}\,}
\newcommand{\musewildlong}{\emph{Multimodal Continuous Emotions in-the-wild sub-challenge}\,}
\newcommand{\musesent}{\textsc{MuSe-Sent\,}}
\newcommand{\musesentlong}{\emph{Multimodal Sentiment in-the-wild Classification sub-challenge}\,}
\newcommand{\musebio}{\textsc{MuSe-Physio\,}}
\newcommand{\musebiolong}{\emph{Multimodal Physiological-Arousal sub-challenge}\,}
\newcommand{\musestress}{\textsc{MuSe-Stress\,}}
\newcommand{\musestresslong}{\emph{Multimodal Emotional Stress sub-challenge}\,}

\newcommand{\awe}{\textsc{RAAW}}
\newcommand{\awelong}{\emph{Rater Aligned Annotation Weighting\,}}

\newcommand{\ds}{\textsc{DeepSpectrum}}
\newcommand{\opensmile}{\textsc{openSMILE}}
\newcommand{\openface}{\textsc{OpenFace\,}}
\newcommand{\egm}{\textsc{eGeMAPS}}

\newcommand{\vgg}{\textsc{VGGish}}
\newcommand{\vggf}{\textsc{VGGface}}
\newcommand{\xce}{\textsc{Xception\,}}

\newcommand{\fau}{\textsc{Facial Action Units\,}}
\newcommand{\mtcnn}{\textsc{MTCNN\,}}
\newcommand{\bert}{\textsc{BERT\,}}

\newcommand{\eg}{e.\,g., }
\newcommand{\ie}{i.\,e., }

\newcommand{\cf}{{cf.\ }}

\copyrightyear{2021}
\acmYear{2021}
\setcopyright{acmlicensed}\acmConference[MuSe '21]{Proceedings of the 2nd Multimodal Sentiment Analysis Challenge}{October 24, 2021}{Virtual Event, China}
\acmBooktitle{Proceedings of the 2nd Multimodal Sentiment Analysis Challenge (MuSe '21), October 24, 2021, Virtual Event, China}
\acmPrice{15.00}
\acmDOI{10.1145/3475957.3484450}
\acmISBN{978-1-4503-8678-4/21/10}
\begin{document}
\fancyhead{}
\title{
The MuSe 2021 Multimodal Sentiment Analysis Challenge: Sentiment, Emotion, Physiological-Emotion, and Stress} % Recognition Classification
% Stress Recognition, BioFANCY
% \aeb{and Physiological-Emotional Recognition of Stress}}

\author{Lukas Stappen}
\affiliation{%
  \institution{University of Augsburg}
  \city{Augsburg, Germany}}

\author{Alice Baird}
\affiliation{%
  \institution{University of Augsburg}
  \city{Augsburg, Germany}}

\author{Lukas Christ}
\affiliation{%
  \institution{University of Augsburg}
  \city{Augsburg, Germany}}

  \author{Lea Schumann}
\affiliation{%
  \institution{University of Augsburg}
  \city{Augsburg, Germany}}
  
\author{Benjamin Sertolli}
\affiliation{%
  \institution{University of Augsburg}
  \city{Augsburg, Germany}}

\author{Eva-Maria Meßner}
\affiliation{%
  \institution{University of Ulm}
  \city{Ulm, Germany}}
  
\author{Erik Cambria}
\affiliation{%
  \institution{Nanyang Technological University}
  \city{Singapore}}

\author{Guoying Zhao}
\affiliation{%
 \institution{University of Oulu}
 \city{Oulu, Finland}
}

\author{Bj\"orn W. Schuller}
\affiliation{%
  \institution{Imperial College London}
  \city{London, United Kingdom}}

\begin{abstract}
\textbf{Mu}ltimodal \textbf{Se}ntiment Analysis (MuSe) 2021 is a challenge focusing on the tasks of sentiment and emotion, as well as physiological-emotion and emotion-based stress recognition through more comprehensively integrating the audio-visual, language, and biological signal modalities. The purpose of MuSe 2021 is to bring together communities from different disciplines; mainly, the audio-visual emotion recognition community (signal-based), the sentiment analysis community (symbol-based), and the health informatics community. We present four distinct sub-challenges: \musewild and \musestress which focus on continuous emotion (valence and arousal) prediction; \musesent, in which participants recognise five classes each for valence and arousal; and \musebio, in which the novel aspect of `physiological-emotion' is to be predicted. For this year's challenge, we utilise the \musesec{} dataset focusing on user-generated reviews and introduce the \ulm{} dataset, which displays people in stressful depositions. This paper also provides detail on the state-of-the-art feature sets extracted from these datasets for utilisation by our baseline model, a Long Short-Term Memory-Recurrent Neural Network. For each sub-challenge, a competitive baseline for participants is set; namely, on test, we report a Concordance Correlation Coefficient (CCC) of $.4616$ CCC for \musewild; $.5088$ CCC for \musestress, and $.4908$ CCC for \musebio. For \musesent an F1 score of $32.82\,\%$ is obtained.
\end{abstract}




\maketitle
%\vspace{-0.3cm}
\section{Introduction}
In the 2nd edition of the \textbf{Mu}ltimodal \textbf{Se}ntiment Analysis in Real-life Media (MuSe) Challenge, we address four tasks incorporating novelties in each: emotion, physiological-emotion, and stress recognition as well as sentiment classification. In the 
\musewildlong (\textbf{\musewild}) and \musesentlong (\textbf{\musesent}), one has to recognise emotional dimensions (arousal, valence) in a regression and classification manner. These tasks are based on work previously outlined for the MuSe 2020 challenge~\cite{stappen2020muse1} %and the corresponding published data set MuSe-CaR, 
and feature substantially improved methods for target creation. The first improvement is the application of \awelong (\awe), a gold standard fusion method for continuous annotations taking both the varied annotator reaction times (aligning) and inter-rater agreements (subjectivity) into account. Additionally, intelligent extraction of valuable features from continuous emotion gold-standards is used to cluster segment-level signals to representative classes so that contributors are faced with two five-way classifications of the level of valence and arousal. These two sub-challenges (\musewild and \musesent) are motivated by the fundamental nature of gold-standard creation on which all tasks and applications of the field are premised.
In the \musestresslong (\textbf{\musestress}), valence and arousal are predicted, from people in stressed dispositions. This sub-challenge is motivated by the high level of stress many people face in modern societies~\cite{can2019stress}. Given the increasing availability of low-resource equipment (\eg smart-watches) able to record biological signals to track wellbeing, we propose the \musebiolong (\textbf{\musebio)}. Adapted from \musestress, the arousal annotations from humans are fused (using \awe) with galvanic skin response (also known as Electrodermal Activity (EDA)) signals for predicting physiological-arousal. Both are set up as regression tasks offering additional biological signals (\eg heart rate, and respiration) for modelling.

For the introduced sub-challenges, two datasets are utilised. As last year~\cite{stappen2020muse1}, we reuse the Multimodal Sentiment Analysis in Car Reviews data (\textbf{\musesec})~\cite{stappen2021multimodal} for the \musewild and \musesent sub-challenges. Including almost 40 hours of video data, it is the most extensive emotion annotated multimodal dataset, gathered in-the-wild with the intention of further understanding real-world Multimodal Sentiment Analysis (MSA), in particular the emotional engagement that takes place during English-speaking product reviews.
%, since it offers around 40 hours of in-the-wild audio-visual recordings which were collected with the intention of gaining further understanding of MSA. 
Within \musesec{}, the subjects are aged between 20 and 60 years, and the spoken word is entirely transcribed. %transcriptions of spoken text and was annotated extensively, including the time-continuous valence and arousal ratings for the entire database. 
For the first time, a sub-set of the novel audio-visual-text \ulmlong dataset (\textbf{\ulm}), featuring German-speaking individuals in a stress-induced situation caused by the Trier Social Stress Test (TSST), is used in this year's \musestress and \musebio sub-challenges. The initial state of \ulm{} consists of 110 individuals (10 hours), richly annotated by self-reported, and continuous dimensional ratings of emotion (valence and arousal). In addition to audio, video, textual features, the \ulm{} includes four biological signals captured at a sampling rate of 1\,kHz; EDA, Electrocardiogram (ECG), Respiration (RESP), and heart rate (BPM). 
Both datasets provide a common testing bed with a held-back labelled test set, to explore the modalities and employ state-of-the-art models under well-defined and strictly comparable conditions.
\begin{table}[t!]
\footnotesize
  \caption{ Reported are the number (\#) of unique videos, and the duration for each sub-challenge hh\,:mm\,:ss. Partitioning of the \musesec{} dataset is applied for each of the two sub-challenges. % The unprocessed duration of the MuSe-CaR dataset is 36\,:52\,:08.
  \ulm{} has a total duration of 5\,:47\,:27 after preprocessing, using the same split for \musestress and \musebio. \label{tab:paritioning}
 }
 \resizebox{\linewidth}{!}{
  \begin{tabular}{l|rrr|rc}
    \toprule
     & \multicolumn{3}{c|}{\textbf{\musesec}} & \multicolumn{2}{c}{\textbf{\ulm}}\\
    Partition & \# & \musewild & \musesent  & \# &  Stress/ Psycho \\
    \midrule
    Train   & 166 & 22\,:16\,:43 &22\,:35\,:55 & 41 & 3\,:25\,:56 \\
    Devel.  &  62 & 06\,:48\,:58 &06\,:49\,:46 & 14 & 1\,:10\,:50  \\
    Test    &  64 & 06\,:02\,:20 &06\.:14\,:08 & 14 & 1\,:10\,:41 \\
    \hline
    $\sum$    & 291 & 35\,:08\,:01 & 35\,:39\,:49 & 69 & 5\,:47\,:27 \\
  \bottomrule
\end{tabular}
}

\vspace{-.3cm}
\end{table}

The goal of the MuSe challenges are to provide a paradigm that is of interest across several communities and to encourage a fusion of disciplines.
% The MuSe challenges goal is to provide both a testing-bed which appeals to several communities as well as encouraging a fusion of disciplines.
%The MuSe challenge goals are to provide a challenge that is of interest across several communities and encourage a fusion of their disciplines. 
We ideally aim for participation that strives for the development of unified approaches applicable to what we perceive as synergistic tasks which have arisen from different academic traditions: on the one hand, we have complex, dimensional emotion annotations that reflect a broad variety of emotions, grounded in the psychological and social sciences relating to the expression of behaviour, and on the other hand, we provide sentiment classes as it is common in sentiment analysis from (multimodal) text-focused modelling.
These fields are rooted within Affective Computing (AC), of which a core aspect is the intelligent processing of uni-modal signals. 
Up to now, the focus in AC when predicting emotion such as by valence and arousal dimensions, was mostly with lower attention to research made on textual information~\cite{ kollias2020analysing,schuller2018interspeech}. %, and if so, usually on spoken language (rather than written such). 
However, the communities appear to be converging even more in recent years (such as supported by the MuSe 2020 \cite{stappen2020muse1} challenge), finding great benefit from multimodal approaches~\cite{arevalo2020gated, gomez2020exploring, qiu2020multimodal}. As an example, both the 2020 and 2021 INTERSPEECH Computational Paralinguistics (ComParE) Challenge have included textual features in an endeavour to more reliably predict valence~\cite{schuller2020interspeech,schuller2021interspeech}.
The second motivation of MuSe is to compare the merits of each of the core modalities (audio, visual, biological, social, and textual signal), as well as various multimodal fusion approaches. Participants can extract their own features or use the provided standard feature sets from the baseline models. 
% This contributes to establishing the extent to which the fusion of approaches is possible and beneficial.
%, as well as advancing emotion recognition systems to be able to deal with fully and semi-naturalistic (in-the-wild) behaviour in large volumes of realistic (user-generated) data. Such data types are the new generation of data utilised for real-world multimedia affect and sentiment analysis~\cite{wang2020discovering} and other research fields~\cite{cuomo2020user}.

% 
% is a challenge-based Workshop focusing on the task of sentiment recognition 
% by means of more comprehensively bridging the audio-visual, language, and further expanding on the previous MuSe challenge by incorporating biological (inc. heart rate and galvanic skin response Electrodermal activity \aeb{(?)}) signals. 

The paper's structure is as follows: First, the four sub-challenges with the corresponding datasets are explained in detail, followed by a description of the challenge conditions. Next, we describe the extracted features from different modalities and the applied pre-processing and alignment for the baseline modelling. Finally, we summarise our baseline results and conclude our findings.
A summary of the challenge results can be found in \cite{stappen2021summary}.

\section{The Four Sub-Challenges}
In the following, we describe and highlight the aforementioned novelties of each sub-challenge, as well as include the guidelines for participation. The evaluation metric for all continuous time-based regression tasks is Concordance Correlation Coefficient (CCC), a well-understood measure~\cite{pandit2019many} of reproducibility, often used in challenges~\cite{valstar2013avec, ringeval2017avec, stappen2020muse1}. The classification task (\musesent) is evaluated in F1 score (macro), a measure robust to class-imbalance. For all challenges with more than one target, the mean of all measures is taken for the final performance evaluation.

\begin{figure}
    \centering
    \includegraphics[width=0.49\columnwidth]{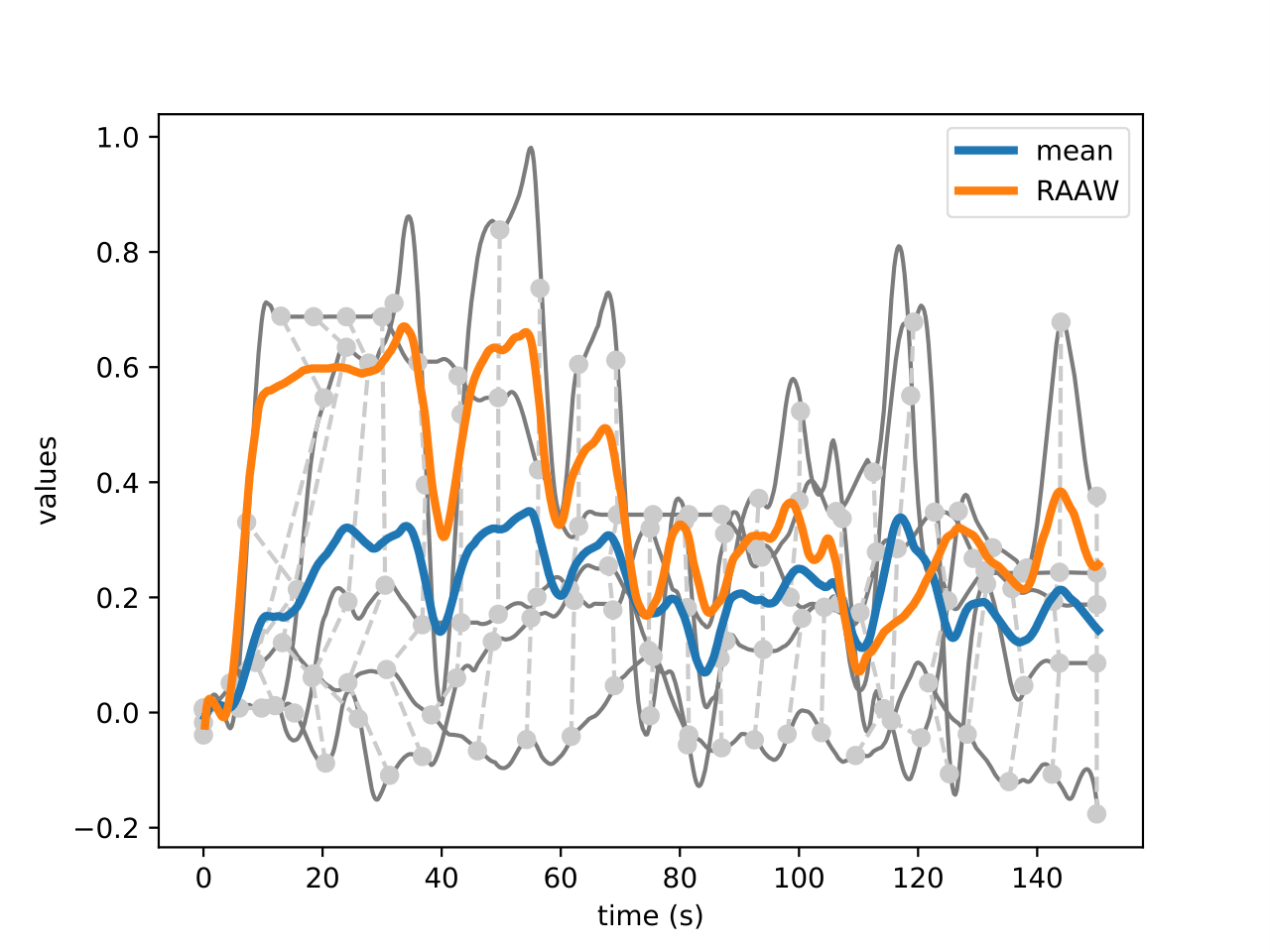}
    \includegraphics[trim={0 0 0 1.4cm},clip,width=0.49\columnwidth]{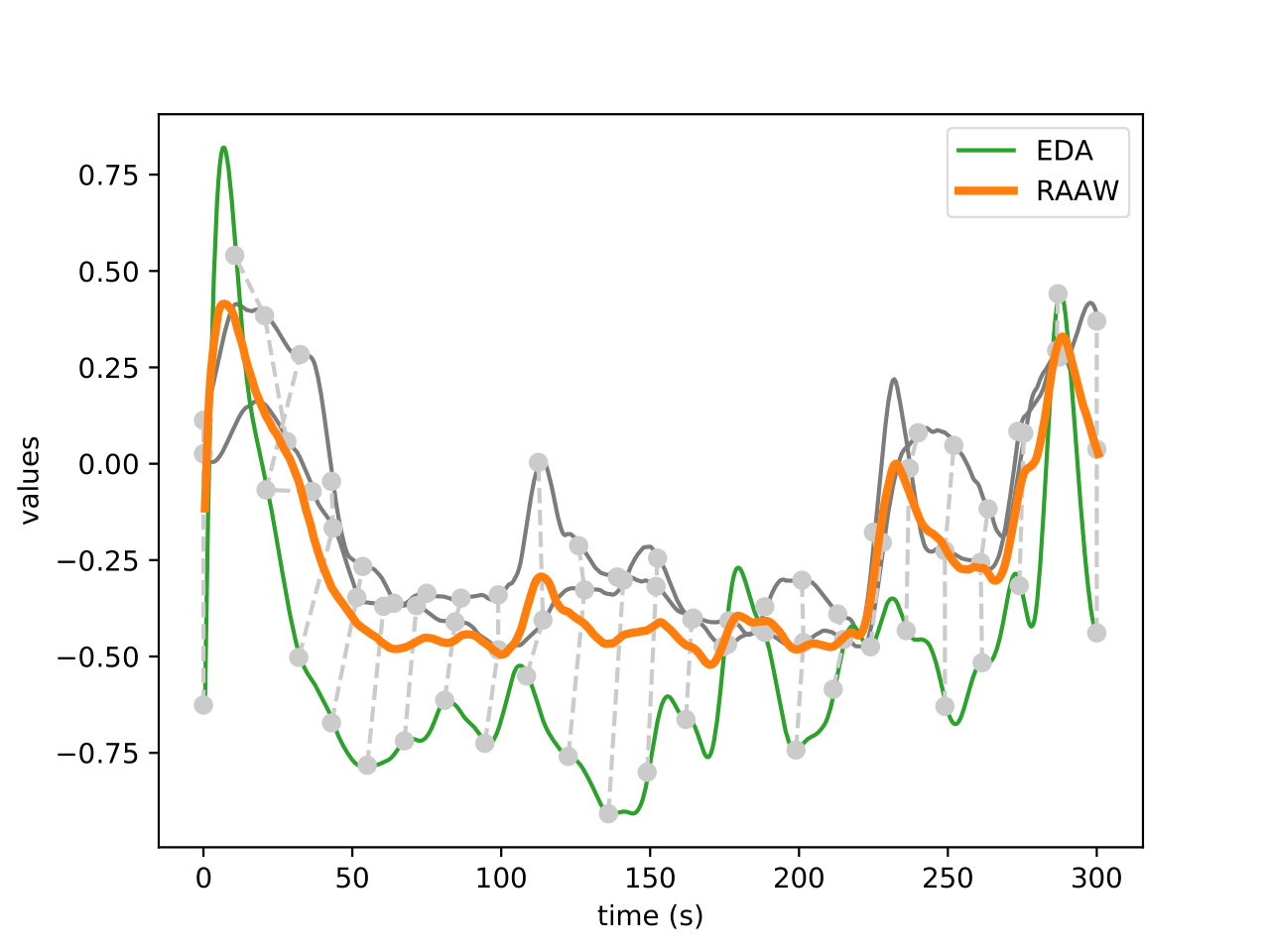}
    \caption{Sample gold standard computation using \awe{} for emotion annotations (dark grey), aligned (warping paths in light grey), and fused from the MuSe-Toolbox~\cite{stappen2021toolbox}. The instance on the left side is from \musewild (id: 100), the resulting signal (orange) is displayed and compared to the average annotation (blue). The right image displays a fusion from \musebio (id:11), where raters 1 and 2 for arousal (dark grey) are combined with the EDA signal (green).}
    \label{fig:awe}
    
\end{figure}

\begin{figure*}
    \centering
    \includegraphics[width=0.19\linewidth]{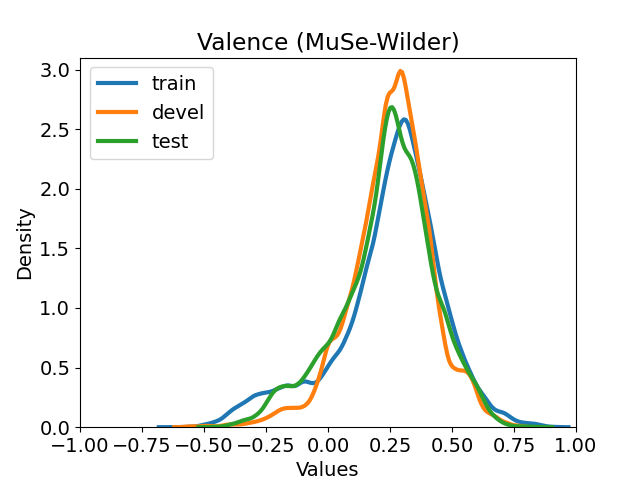} 
    \includegraphics[width=0.19\linewidth]{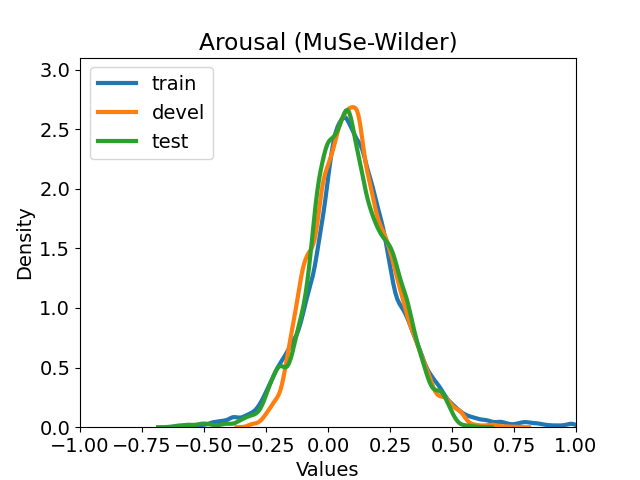} 
    \includegraphics[width=0.19\linewidth]{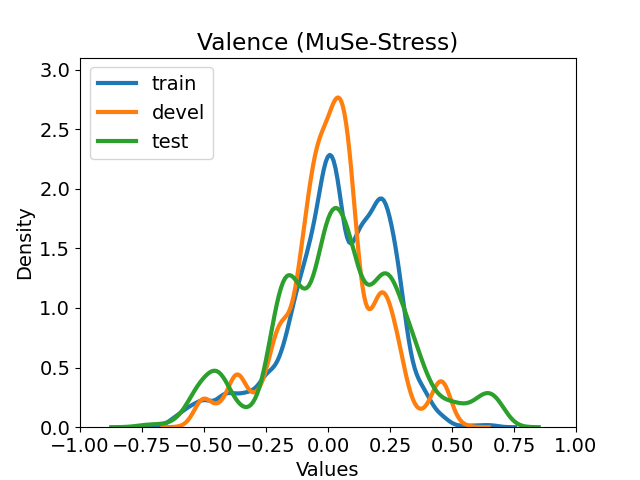}
    \includegraphics[width=0.19\linewidth]{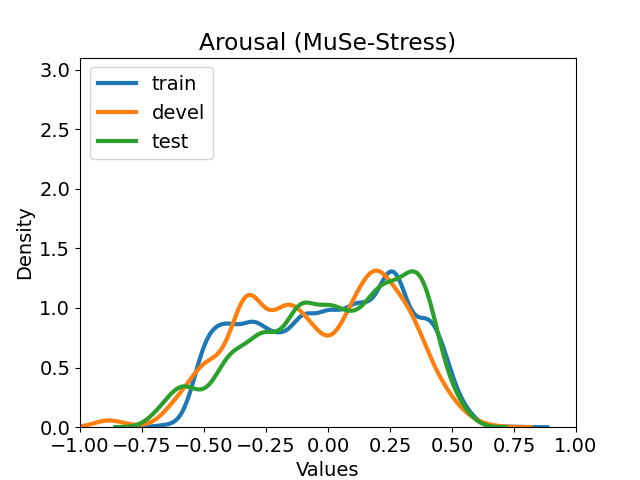}
    \includegraphics[width=0.19\linewidth]{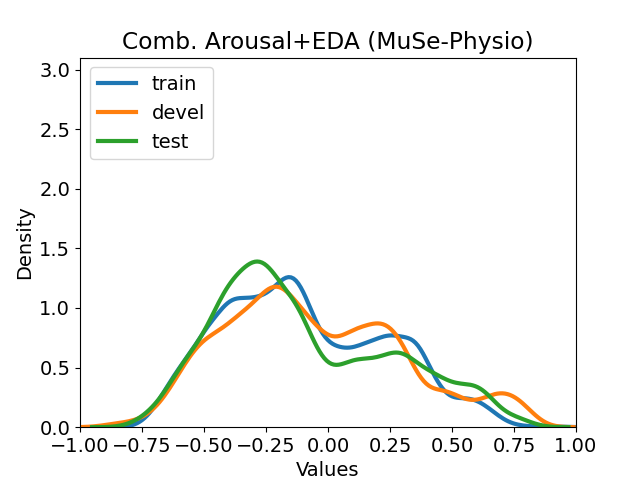}\\
    %\vspace{-0.2cm}
    \caption{Frequency distribution in the partitions train, (devel)opment, and test for the continuous prediction sub-challenges \musewild, \musestress, and \musebio. For all sub-challenges, the value distributions between the partitions are fairly similar.}
    \label{fig:freq}
    %\vspace{-0.4cm}
\end{figure*}

\subsection{The \musewild Sub-Challenge\label{sec:wild}} 
The \musewild is an extension of the MuSe-Wild 2020 sub-challenge, where participants had to predict emotional dimensions (valence, arousal) in a time-continuous manner. The amount of data utilised from \musesec{} is shown in \Cref{tab:paritioning}. The valence dimension is often referred to as the emotional component of the generic term sentiment analysis and is often used interchangeably~\cite{thelwall2010sentiment, mohammad2016sentiment,preoctiuc2016modelling}. Human annotation of continuous emotions leads to disagreements between raters, \eg due to differences in perception~\cite{hoffmann2012mapping} and reaction time~\cite{nicolaou2014dynamic}, which should be mitigated by fusion to a gold standard. Since this signal is the prediction target, a variety of fusion methods are available in the literature~\cite{panagakis2015robust, grimm2005evaluation} and this development has motivated other challenges~\cite{ringeval2018avec}.

This year's \musewild emotion recognition task is based on a completely novel continuous annotator fusion technique \awe, which targets the difficulties of combining subjective emotion annotations for a gold standard annotation present. For this, we employ our fusion method on a minimum of five different ratings that weights inter-rater agreements as well as considers the varied reaction times as displayed in \Cref{fig:awe}. The varying rater lag that is inherent to all annotation signals will be targeted by aligning the standardised (per annotator) ratings using a generalised Canonical Time Warping (CTW) method~\cite{zhou2015generalized}. The Evaluator Weighted Estimator (EWE)~\cite{grimm2005evaluation} is then used to fuse the aligned, individual signals by weighting a signal depending on the inter-rater agreement to the mean of all others. This technique is described in length in \cite{stappen2021toolbox}.
% \iffinal{~\cite{Toolboxpaper}}. 
The resulting distribution is shown in \Cref{fig:freq}.
% The \musewild sub-challenge uses the \musesec database

\subsection{The \musesent Sub-challenge}
%Table: Cluster names + conclusion of features}
%The first group -- rooted in the field of Sentiment (and Opinion) Mining and specialising in Natural Language Processing (NLP) methods for symbolic information analysis -- leverages the text modality, and focuses on the prediction only of discrete sentiment label categories~\cite{zadeh2018proceedings}. 
Mapping continuous emotion annotations to discrete classes are considered a highly ambiguous and challenging task and have so far hardly been computed successfully in a time-continuous fashion~\cite{wollmer2008abandoning}. In general, classes are often considered a simplified concept for interpretation compared to dimensional representations. In \musesent, participants will have to predict five advanced sentiment classes for each emotion dimension of valence and arousal on a segment-level, based on audio-visual recordings and the transcribed speech of \musesec. The sub-challenge uses the topic-based segmentation from MuSe 2020~\cite{stappen2020muse1}. The classes are extracted using a novel method of the MuSe-Toolbox~\cite{stappen2021toolbox}.
%\iffinal{~\cite{Toolboxpaper}} 
which aims to find a mapping between continuous dimensional and categorical representations of emotion through the extraction of time-series features and the application of unsupervised clustering.

More specifically, we first extract a range of time-series features on a segment-level\footnote{arousal: median, standard deviation, percentile \{10, 90\}, relative energy, relative sum of changes, relative number of peaks, relative longest strike \{below, above\} mean, and relative count below mean; Valence: the same features as for arousal, and additionally: mean, percentile \{5, 25, 33, 66, 75, 95\}, and the percentage of reoccurring data-points to all data-points} based on the continuous \awe-fused annotations. The absolute features are normalised depending on the varying length of a segment to limit undesirable properties solely due to the influence of the segment length. 
To reduce the feature space, we apply Principal Component Analysis (PCA) to project our data to a five dimensional space of principal components which are derived from the eigenvectors of the covariance matrix.
The transformed data is further clustered into five class clusters using a) the $k$-means algorithm~\cite{lloyd1982least} for valence and b) a Gaussian Mixture clustering model%(apply a full co-variance matrix
~\cite{duda1973pattern} for arousal.
%Finally, features are transformed to five clusters by a Fuzzy C-Means (FCM) algorithm~\cite{Mingoti:33} where $m$ is set to $2$, as proposed in~\cite{Hathaway:34}.
%For FCM the number of iterations was capped at $10$\,k and the error threshold for early stopping was set to $50^-4$. %, and we used the random seed $301$ for reproducibility.
To ensure that the development and test set have no effect on the generated classes, we only apply this process on the training set segments. The segments belonging to the development and test partitions, are then `predicted' by assigning the cluster with the closest centre to each data-point. 
These clusters are evaluated through both qualitative and quantitative measures: \begin{inparaenum}[(1)] \item the amount of data-points of the smallest class is larger than a quarter of by-chance-level\footnote{For example, five classes resemble a by-chance level of 20\,\%, thus, the smallest class have to cover at least 5\,\% of the data points} \item to evaluate cluster cohesion and separation, the widely used Silhouette Coefficient (SILC)~\cite{rousseeuw1987silhouettes} is calculated, ranging from -1 to 1 (closer to 1 is superior).
%, and the \textit{Fuzzy Partition Coefficient} (FPC)~\cite{xie1991validity}, specifically applicable for non-crisp clustering, ranging from 0 to 1 (1 represents perfect clustering).
\end{inparaenum}For the two chosen settings, we achieve a SILC of $0.19$ and $0.10$, respectively for valence and arousal clusters. The PCA leads to a denser representation along the orthogonal axes, making a higher SILC value hard to achieve, since the metric is prone to error when clusters show different kinds of cluster densities~\cite{liu2010understanding}, which naturally occurs in this setting. 
% hardly distinguishable clusters and is especially sensible towards densities
%This is considered to be a reasonable result for real-world data clustering. 
%  A minor disadvantage of the SILC is that it cannot distinguish subclusters well and is also affected by various densities
%\miss{this can be explained... }\cite{liu2010understanding}.

\begin{figure*}[t!]
    \centering % to be replaced
    \includegraphics[ width=.7\columnwidth]{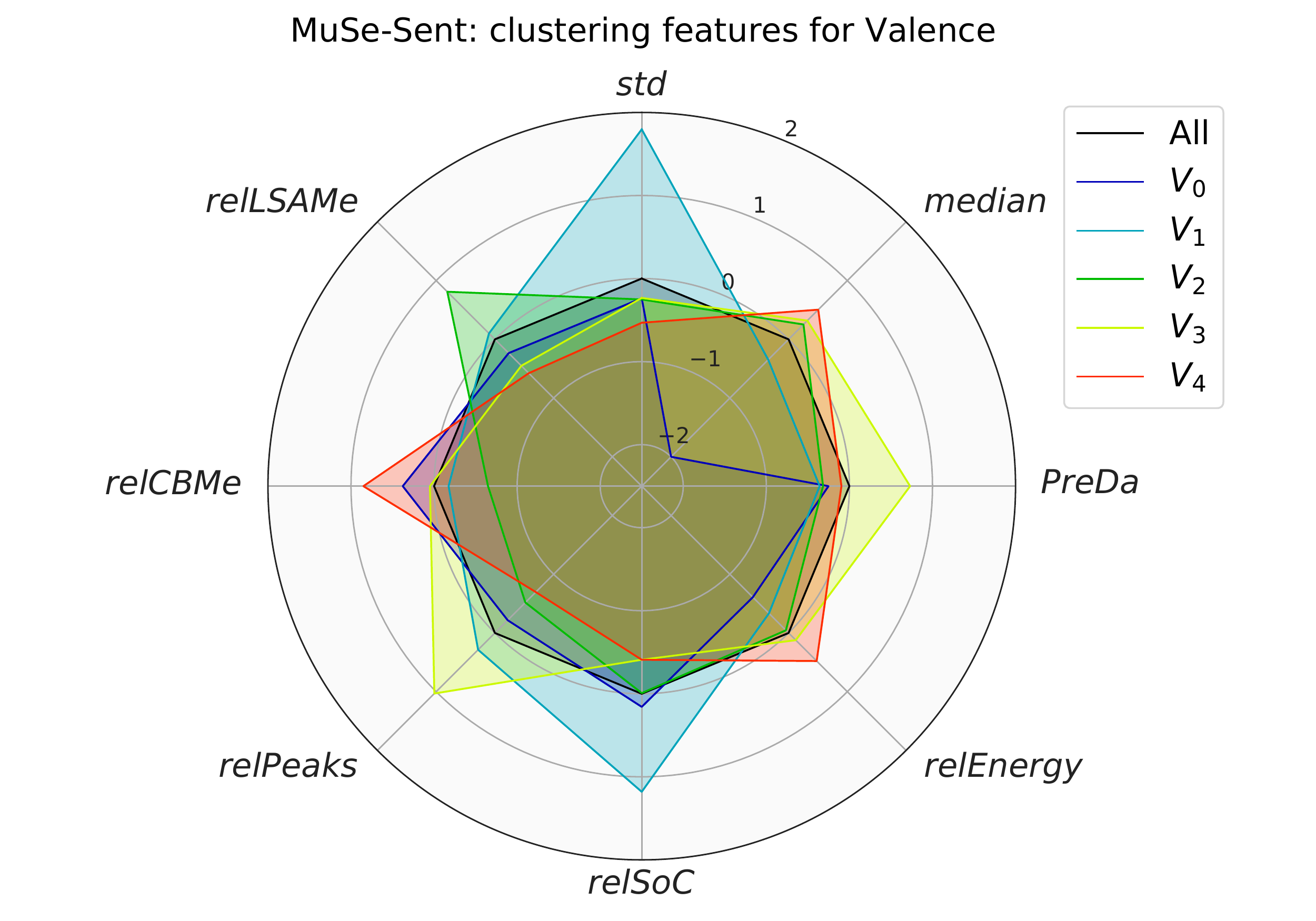}
    \includegraphics[ width=.7\columnwidth]{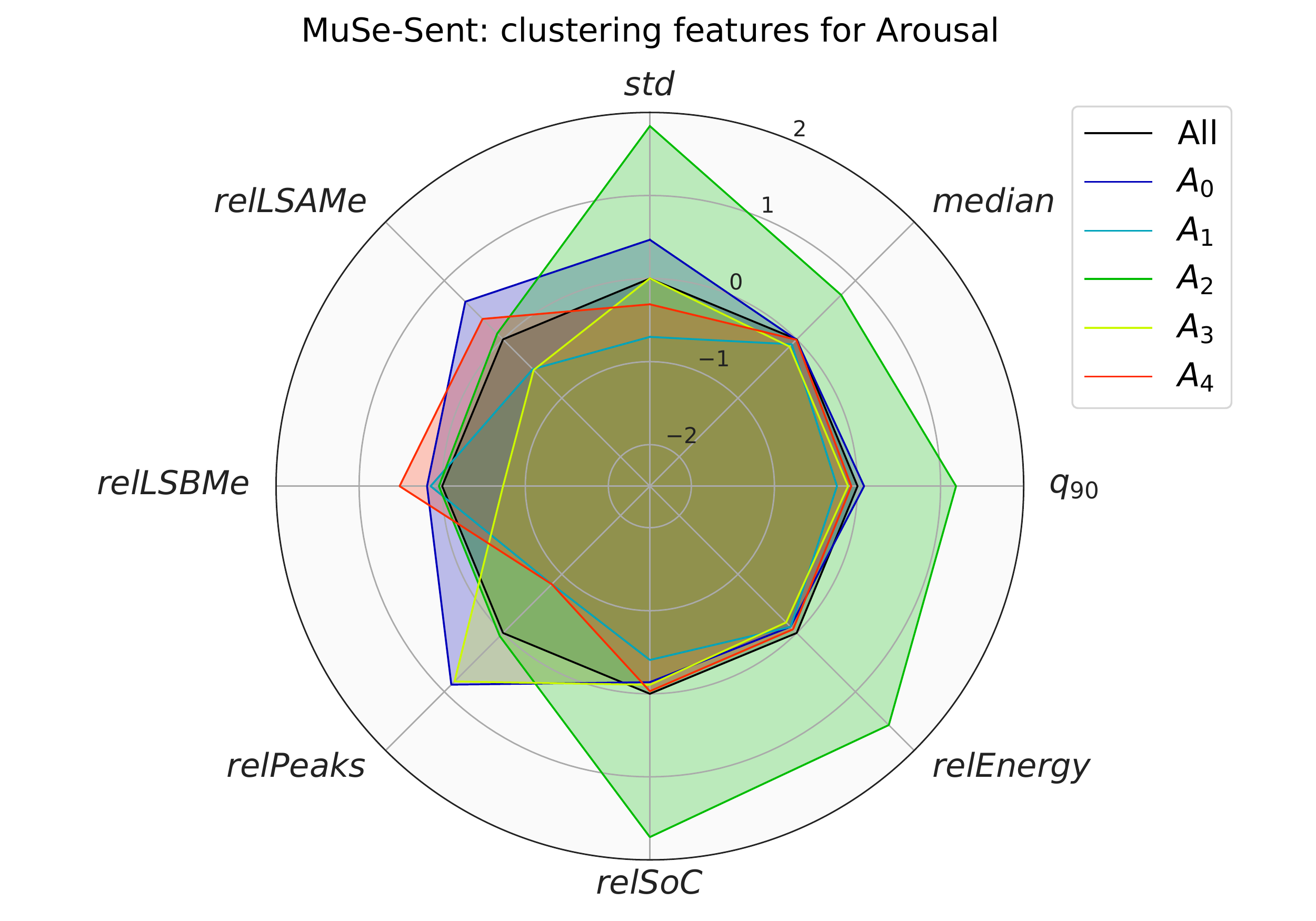}
    \caption{
    Mean of selected clustering features for each of the created classes which are used in the \musesent sub-challenge. The features shown are standard deviation ($std$), $median$, the 90th percentile ($q_{90}$), percentage of reoccurring datapoints to all datapoints ($PreDa$), relative energy ($relEnergy$), relative sum of changes ($relSoC$), relative number of peaks ($relPeaks$), relative count below mean ($relCBMe$), relative longest strike below mean ($relLSBMe$), and relative longest strike above mean ($relLSAMe$).
    Features indicated by ``relative'' ($rel$) are normalised by segment length. Additionally, all features are standardised, hence, the mean value of all data is always equal to zero. Illustrations generated by the MuSe-Toolbox~\cite{stappen2021toolbox}}
    \label{fig:sent-cluster-features}
\end{figure*}

Since the features reflect characteristics of the emotional annotation and not just the mean value as in last years' MuSe-Topic task~\cite{stappen2020muse1}, class descriptions, \ie low, medium, or high would inadequately reflect the meaning. % Mapping continuous emotion signals to discrete classes is considered an incredibly difficult task due to a lack of understanding of the created (in an unsupervised manner) classes and has been rarely computed successfully~\cite{wollmer2008abandoning}. 
With this in mind, to gain understanding of the classes, we %further analyse statistical measures, correlation between the classes, and the segment features originally used as the clustering input, as well as the most
display the most distinctive features in \Cref{fig:sent-cluster-features} for interpretation showing the named valence classes as $V_{\#}$ and arousal as $A_{\#}$ while $\#$ represents the class number, not implying any specific order. For example, segments from classes $V_1$ and $A_2$ have a comparatively large (to the mean of all other classes) `standard deviation' and `sum of changes', which indicates a higher annotation fluctuation and intensity than other classes. The distribution of segments across the classes can be found in \Cref{tab:class_distr}.

% metrics % moved this to the top of all challenge introduction 
% As for measuring classification performance, the macro-averaged F1 score (Macro F1) is calculated independently for valence and arousal predictions. The mean of the valence and arousal score (combined) is then used as the final performance assessment. 

\begin{table}[t!]
\caption{Distribution of the valence and arousal classes across partitions used in the \musesent sub-challenge as a result of our configured class search from the MuSe-Toolbox~\cite{stappen2021toolbox}.}
\resizebox{0.7\linewidth}{!}{
    \begin{tabular}{@{}lrrr||lrrr@{}}
    \toprule
    \multicolumn{4}{c}{\textbf{Valence}}  &  \multicolumn{4}{c}{\textbf{Arousal}}\\ 
    \midrule
               & \multicolumn{1}{l}{Train.} & \multicolumn{1}{l}{Devel.} & \multicolumn{1}{l}{Test} & & \multicolumn{1}{l}{Train.} & \multicolumn{1}{l}{Devel.} & \multicolumn{1}{l}{Test} \\ 
               \midrule
    0 & 528                     & 71                        & 89 & 0 & 612                      & 249                       & 178 \\
    1 & 552                     & 159                       & 277 & 1 & 534                      & 135                       & 194 \\
    2 & 1178                    & 458                       & 378 & 2 & 312                      & 96                       & 53 \\
    3 & 1112                    & 405                       & 271 & 3 & 1255                     & 388                       & 448 \\
    4 & 837                     & 242                       & 245 & 4 & 1494                     & 467                       & 387 \\
    \midrule
    $\sum$      & 4207                      & 1335                      & 1260 & $\sum$ & 4207                      & 1335                      &   1260    \\
    \bottomrule
    \end{tabular}
}
%\vspace{-1cm}
\label{tab:class_distr}
\end{table}

\subsection{The \musestress Sub-challenge}
In the \musestress, participants will have to predict valence and arousal in a time-continuous manner. This sub-challenge is motivated by real-world applications for emotion recognition and further motivated by stress in modern life. In this novel sub-challenge, the idea of `multimodal' sentiment analysis is pushed further by the inclusion of biological signals that have been shown to be applicable for recognising physiological stress~\cite{pourmohammadi2020stress}, and for emotion recognition~\cite{shukla2019feature}. 

Participants are provided with the multimodal \ulm{} database, in which subjects were recorded under a highly stress-induced free speech task, following the TSST protocol~\cite{kirschbaum1993trier}. For the TSST, after a brief period of preparation the subjects are asked to give an oral presentation, within a job-interview setting, observed by two interviewees who remain silent for the period of five minutes.
To allow consistent data partitions, we only keep data recorded under the same experimental conditions. The resulting 69 participants (49 of them female) are aged between 18 and 39 years, providing a total amount of about 6 hours of data for the \musestress and \musebio sub-challenges (\cf \Cref{tab:paritioning}). Besides audio, video, and text, the participants can optionally utilise the ECG, RESP and BPM signals. 

The dataset has been rated by three annotators continuously for the emotional dimensions of valence and arousal, at a 2Hz sampling rate, and a gold standard is obtained by the fusion of annotator ratings, utilising the \awe{} method, as described in \Cref{sec:wild} from the MuSe-Toolbox~\cite{stappen2021toolbox}. When creating the fusion a mean CC inter-rater agreement of 0.204 ($\pm$ 0.200) for valence and  0.186 ($\pm$ 0.230) for arousal is obtained.
The distributions of the valence and arousal signals for the dataset are depicted in \Cref{fig:freq}. 

% For this, we also utilise \awe{} 
% \iffinal{~\cite{MuSeToolbox}}. 

\subsection{The \musebio Sub-challenge}
In the cross-modal \musebio, participants will have to predict a combined signal of arousal and EDA. Again, for this task, the \ulm{} dataset is employed, where the TSST was utilised as a standardised and renowned stressor, allowing for a controlled setting with high-quality data while maintaining a naturalistic subject behaviour.

Physiological signals, including EDA have been used as a feature in at least one other multimodal emotion challenge~\cite{dhall2020emotiw}. However, we consider this sub-challenge to be the first time the physiological signal is combined with the emotional -- human-annotated -- signal. From all the biological signals available in the \ulm{} dataset, we choose to use the EDA signal, as not only are the signal characteristics subjectively similar to continuous emotion as exemplarily depicted in \Cref{fig:awe}, but the signal itself has been shown in the literature to be a psycho-physiological indication of emotional arousal~\cite{caruelle2019use}. Given that in the context of an interview, arousal may also appear to be a more hidden emotion, we consider that the fusion of arousal and EDA may improve recognition and offers a more objective marker for a speaker's arousal~\cite{caruelle2019use}. Further variants are introduced in~\cite{baird2021physiologically}.
To obtain the combined emotion and EDA signal gold standard, we again utilise the \awe{} fusion strategy. However, in this case, the lowest weighted annotator is excluded and replaced with the EDA signal. The EDA signal is downsampled to 2\,Hz and smoothed slightly before fusion through a Savitzky–Golay filtering approach (window size of 26 steps), to avoid irrelevant, fine-grained artefacts in the signal. For this gold standard of emotion, we obtain an inter agreement of 0.233 ($\pm 0.289$), which was improved compared to the arousal gold standard obtained in \musestress.
% Therefore in the case where inter rater agreement between annotations is low, we propose that replacing this with an available biological signal may show promise.

\subsection{Challenge Protocol}
As part of the mandatory prerequisites required to play a part in the challenge, interested participants are obliged to download and fill in the End User License Agreement (EULA) which is put forward through the homepage\footnote{\url{https://www.muse-challenge.org/muse2021}}. On top of this, participants are further required to hold an academic affiliation.  
Each participation must be accompanied by a paper (6-8 pages in length including references) reporting the results obtained and methods applied. The organisers also consider general contributions in the field. 
%BS: This is a lot of bla which I cut:
%To ensure papers that qualify for the challenge match with the required workshop criteria, a technical program committee is delegated carrying out a 
Peer review is double-blind. %peer-review process for the submitted papers put forth. If accepted, the contribution will be part of the proceedings. 
To obtain results on the test set, the participants can upload their predictions up to five times per 
%BS: Please UNIFY! Sub-Challenge or sub-challenge (appears to be the regular case)
sub-challenge, whose labels are unknown to them. We want to point out that the organisers only evaluate the participants' results but do not participate themselves as competitors in the challenge. 
% For contenders aspiring to be featured temporarily on the public leader board on the MuSe homepage, should further provide an arXiv preliminary technical report link, and a Github repository uploaded with their source code. However, a detailed description of methods and results together with a citation of this paper should go along with all entries. 

%A \emph{gold-standard} was computed on the individual annotators using an Evaluator Weighted Estimator (EWE) approach, in which inter-rater agreement is considered. is described, \eg further in~\cite{schuller2013intelligent} and has been applied to similar continuous emotion-based tasks~\cite{ringeval2017avec}, and corpora~\cite{ringeval2013introducing}. 

\section{Baseline Features and Model}
To save effort and time which would be incurred by the participants while extracting various features from the large datasets provided, we put forth a selection of features drawn from the video data for each sub-challenge. In a more elaborate outline, the available features comprise of seven model-ready video, audio, and linguistic feature sets\footnote{Note: Furthermore, we place it at the applicants' disposal to use (unaligned) features from MuSe2020 for \musewild and \musesent as well as external datasets and features under the condition that this should be clearly explained in their accompanying paper. These sources could be, \eg commercial or academic feature extractors, libraries, or pre-trained networks.}. The amalgamation of features provided surpasses most other related audio-visual challenges~\cite{challengehml2020grand, dhall2020emotiw, egede2020FG}.
% valstar2013avec, ringeval2017avec, kollias2019deep, zadeh2018proceedings}. 
In respect to the annotation sampling rate, the features are extracted at a step size of 0.25\,s for the \musesec{} and 0.5\,s for the \ulm{} dataset.
%The features are to be delineated following each modality in the adjacent section. 

\subsection{Pre-processing}
The data of both datasets has been partitioned into a Train, Development, and Test partition. Emotional ratings, speaker independence, and duration are considered when creating the partitions (\cf \Cref{tab:paritioning} for an overview). Since the amount of recordings made available between sub-challenges can vary, so too does the time required to extract the most applicable features during the pre-processing stages. Aiming to minimise the distortion of the task objectives, we deliberately omit advertisement sections of the videos for the \musesec-based sub-challenges. In the \ulm{} dataset, each video is cut to exclude scenes outside of the TSST setting, \eg excluding participants' names. For both datasets, the segments are crafted with the focus on the active voice based on the sentence transcriptions or if a visible face applies. For \musesent, we adjacent segments in instances where the segments deals with the same topic and the gap is less than two seconds.

%\vspace{-0.2cm}
\subsection{Acoustic}
\opensmile{} and \ds{} are well-established tools for the extraction of acoustic emotional feature representations. Most notably, they have proved valuable in the extraction of audio processing tasks in renowned challenges in speech emotion recognition (SER)~\cite{schuller2020interspeech,schuller2021interspeech}.
For all acoustic features, a six second window size is applied. In the first step of the pre-processing pipeline, the full audio is extracted from a given video. The second step is the conversion of the audio from stereo to mono to 16\,kHz, 16\,bit after its normalisation to -3 decibels.

\subsubsection{\egm}
The prevalent open-source \opensmile{} toolkit~\cite{eyben2010opensmile} is used to extract the extended Geneva Minimalistic Acoustic Parameter Set (\egm)~\cite{eyben2015geneva}. Comprising of 88 acoustic parameters for automatic voice analysis tasks~\cite{stappen2019speech}, it is a minimal set of hand-crafted features relying on affective physiological changes in voice production that has previously proven valuable for a variety of emotion research~\cite{baird2019can,stappen2020cross,stappen2020muse1}. %shown to be very effective 

% Furthermore, low-level brute-force descriptors of 130 dimensions which are computed with \opensmile{} are also provided. They comprise of deltas and double-deltas commonly featured as 1st and 2nd order derivatives of the audio signal. In closing, a window size of 10 ms is carried forth on the default LLD extraction  configuration.

\subsubsection{\ds}
The prime function of \ds~\cite{amiriparian2017snore} is to utilise the spectral features acquired from speech instances within a pre-trained image recognition Convolutional Neural Networks (CNNs). The consecutive inputs result in the extraction of feature vectors. A commonly applied architecture in this framework is VGG-19~\cite{simonyan2014very}. Here, we keep the default settings for extraction to obtain a 4\,096 dimensional feature set. 

\subsubsection{\vgg}
In addition, we extract \vgg{} functions~\cite{hershey2017cnn} pre-trained on an extensive YouTube audio dataset (AudioSet)~\cite{gemmeke2017audio}. The underlying data contains 600 classes, and the recordings contain a variety of `in-the-wild' noises that we expect to be beneficial to obtain robust features from our `in-the-wild' videos. By aligning the frame and hop size to the annotation sample rate, we extract a 128-dimensional \vgg{} embedding vector every 0.25\,s from the underlying log spectrograms.

%\vspace{-0.2cm}
\subsection{Vision}
Extracting specific image descriptors that match certain attributes, \eg face, remains the paramount focus of most visual feature extractors. Our offered visual feature sets are inclined to capture the entire surroundings
as well as analysing human behaviour synthesised from gesture and facial expressions. Participants are also provided with an array of extracted faces which are directly extracted from the raw frames. 

%\vspace{-0.15cm}

\subsubsection{\mtcnn}
% Its functionalism is supported by a three-phase configuration that pulls out landmarks and previses any facial expression. 
The \mtcnn~\cite{zhang2016mtcnn} is used to distinguish facial expressions captured in the videos, pretrained on the data sets WIDER FACE~\cite{DBLP:journals/corr/YangLLT15b} and CelebA~\cite{liu2015faceattributes}. 
%It also provides a confidence measure that enables the calibration of the false-negative and false-positive rate.   %However, we agreed not to modulate the confidence threshold as the facial extraction structures do not provide features for false positives. 
For \musesec, we examined the extraction as described in detail in~\cite{stappen2020muse1}, where an F1 score of $86$\,\% on a labelled subset was achieved. Compared to these highly dynamic camera positionings (zoom, free etc.), \ulm{} has a static setting. In an visual inspection aimed to control the performance, an apparently flawless extraction was found. The extractions were ultimately put in use as inputs for \vggf{} and \openface. 

% The identified bounding boxes were classified into false and true positives putting into consideration the overlap of intersection (IoU). Subsequently, the IoU was calculated based on videos elected from several channels in a quantitative performance analysis. An accuracy of $90$\,\%, from the detector was achieved in conjunction with an F1 score of $86$\,\% on the \musesec selection. 
% as the \mtcnn practical capabilities had been verified by both outcomes. 
% \td{@Alice: How do we justify it for Stress?}

%\vspace{-0.15cm}
\subsubsection{\vggf}
\vggf{} (version 1)~\cite{Omkar2015recognition} is aimed at the extraction of general facial features for images obtained by \mtcnn in cropped versions. The visual geometry group of Oxford introduced the deep CNN referred to as VGG16~\cite{simonyan2014very}. The training data constitutes of 2.6 million faces and over 2\,500 identities. The \vggf{} architecture was originally intended for supervised facial recognition purposes~\cite{simonyan2014very}. However, detaching the top-layer of a pretrained version results in a 512 feature vector output referred to as \vggf. Presenting high levels of performance while consuming less data is the  main advantage held for \vggf{} in comparison to other facial recognition models. % These features can also be used to learn predictive facial movements.

%\vspace{-0.15cm}
\subsubsection{OpenFace}
Facial features in 2D (136 dimensions) and 3D (204 dimensions), gaze positions (288 dimensions), intensity and activity of 17 Facial Action Units (FAUs) for both center and left side, and 6 head stances were extracted from cropped faces identified using \mtcnn. This was achieved through the wide array of facial features offered by the  \openface~\cite{Baltruvsaitis16OAO} toolkit. For the \ulm data challenge, we only provide intensity, as activity features appear to be of less use for this task.

\subsubsection{\xce}
Generally used to extrapolate generic vision features, \xce~\cite{he2016deep} should provide participants with environmental features using stacked residual blocks\footnote{not used for \musestress or \musebio as recording environment for \ulm{} changes only minimally and participants showed minimal movement due to their stressful situation.}. Among other challenges, it came in first on the ILSVRC 2015 classification challenge. The network is pre-trained on the ImageNet dataset compromising of 350 million images and 17\,000 classes. The then frozen network architecture prepossesses a given frame through the layers until the last fully connected layer from which a 2\,048 deep feature dimensional vector is obtained.

\subsection{Language: Bert}
%~\cite{bojanowski2017enriching} is based on the skip-gram model, where the representation of a vector accompanies each n-gram word. The word embedding library was trained on the English Common Crawl corpus (600B tokens). Unlike other word embeddings models word2vec~\cite{mikolov2013distributed}, and GloVe~\cite{pennington2014glove}, the available sub-word chunks make it more likely to compute word mappings for any word (hence unknown) that was not part of the original training corpus. This functionality proves useful as it allows us to transform 96\,\% word embedding vectors from words.
 
%\subsubsection{\ft}
%\rep{~\cite{bojanowski2017enriching} is a library for efficient learning of word embeddings. It is based on the skipgram model where a vector representation is associated to each character n-gram. The model is trained on the English Common Crawl corpus (600B tokens). In comparison to other traditional word embeddings, such as, word2vec~\cite{mikolov2013distributed}, or GloVe~\cite{pennington2014glove}, these sub-words chunks make it possible to calculate word representations of words which were not part of the original training corpus (out-of-vocabulary).This appears advantageous since we work with a domain-specific corpus including technical terms and model names. This a valuable function, and enables us to transform 96\,\% of words to word embedding vectors.} %In~\cite{joulin2017bag} they propose a simple and efficient baseline for text classification. Even though it is faster for training and evaluation, it shows competitive performance.

The text feature extraction process employs a Transformer language model, namely Bidirectional Encoder Representations from Transformers (\bert)~\cite{devlin2019bert}, which have already been successfully used for a variety of NLP tasks~\cite{schuller2021interspeech,schuller2020interspeech,stappen2020cross, stappen2020summary,stappen2020uncertainty}. \bert pre-trains its deep representations on context of unlabelled text before fine-tuning them on a broad selection of down-streaming NLP tasks. During inference, the context-based representations are preserved, excerpting one vector per word. This is in contrast to static word embeddings which give one vector per word independent of the context. Our features are the sum of the last four \bert layers resulting in a 768 dimensional feature vector analogous to~\cite{sun2020multi}. For \musewild and \musesent, the base variant of \bert, pretrained on English texts, is used. Analogously, as the \ulm{} data set is in German, for \musestress and \musebio, the \bert (base) pretrained on German texts is utilised.

%\vspace{-0.3em}
\subsection{Alignment}
%\subsubsection{Word-Transcription Alignment}

The extensive assortment of features are from three modalities. The corresponding sampling rate of each modality differs, which leads to a different length of the extracted features along the time axis. All visual features
are incessant through the video with a frame sampling of 4\,Hz for \musesec{} and 2\,Hz for \ulm{}, which is equivalent to the labelling rate. The audio sampling of \ds, and that of \egm{} apply the same frequency. \vgg{} and \fau are the only feature sets relying only on frames where a face is observable. By the nature of text, the corresponding features do not follow a fixed sampling rate, as the duration of a spoken word varies. 

For each sub-challenge, we make label-aligned features available. These have accurately the same stretch and time-stamps as the provided label files. We apply zero-padding to the frames, where the feature type is absent. Such instances include \textsc{\openface}, when no face appears or extraction fails, \eg when only small faces appear in the original frame. 
The text features are repeated for the interval of a word and non-linguistic parts are also imputed with zero vectors. \musesec{} offers automatic, word-aligned transcriptions \cite{stappen2021multimodal}. For \ulm, manual transcripts of the videos are available. We use the Montreal Forced Aligner (MFA)~\cite{mcauliffe2017montreal} tool to obtain time-stamps on the word level. The MFA includes pretrained acoustic models, grapheme-to-phoneme models, and pronunciation dictionaries for various languages. We use the German (Prosodylab) model and the German Prosodylab dictionary to align the \ulm{} transcripts. The time-stamps yielded by the MFA are used to align the word embeddings to the 2\,Hz frames in the \ulm{} dataset. 
% Moreover, they are provided to the participants.
% This groundwork should enable the participants to get started in the shortest time frame while giving room to extract and align own features. If sought after, audio-video features extracted by the participants can be aligned using the corresponding timestamps from the label files. 

\subsection{Baseline Model: LSTM-RNN\label{sec:model}}
In order to address the sequential nature of the input features, we utilise a Long Short-Term Memory (LSTM)-RNN based architecture. The input feature sequences are input into uni- and bi-directional LSTM-RNNs with a hidden state dimensionality of $h = \{32, 64, 128\}$, to encode the feature vector sequences. We test different numbers of LSTM-RNN layers $n = \{1, 2, 4\}$. Based on experiences from initial experiments, some hyperparameter searches are task-dependently executed:
\musewild we search for a suitable learning rate $lr=\{0.0001, 0.001, 0.005\}$; for \textit{\musesent} $lr=\{0.001, 0.005, 0.01\}$; for \textit{\musestress} and \textit{\musebio}  $lr=\{0.0001, 0.0002, 0.0005, 0.001\}$. As we observed overfitting in some settings of  \musebio, we also tried L2-Regularisation with a penalty of $0.01$ for this task.

The sequence of hidden vectors from the final LSTM-RNN layer is further encoded by a feed-forward layer that outputs either a one-dimensional prediction sequence of logits for each time step (regression), or a single-value per prediction target (classification).

In the training processes, the features and labels of every input video are further segmented via a windowing approach \cite{sun2020multi, stappen2020muse1,stappen2021multimodal}. For \musewild and \musesent, we use a window size of 200 steps (50 s) and a hop size of 100 steps (25 s). For \musestress and \musebio, a window size of 300 steps (150 s) and a hop size of 50 steps (25 s) proved to be reasonable choices.

\subsection{Fusion}
We apply %feature-level (early) and 
decision-level (late) fusion to evaluate co-dependencies of the modalities.
%Feature-level fusion focuses on early state interactions by combining the raw features directly before feeding them as a concatenated input vector into the described architecture. 
The experiments are restricted to the best performing features from each modality only. For decision-level fusion, separate models are trained individually for each modality. The predictions of these are fused by training an additional LSTM-RNN model as described above. For all continuous regression tasks, we apply uni-directional version with $lr = 0.0001$, $h = 64$, and $n = 1$, and for \musesent a bi-directional one with $lr = 0.005$, $h = 32$, and $n = 2$.

%\vspace{-0.1cm}
\section{Experiments and Baseline Results}

For all sub-challenges, the same network architecture is applied (\cf \Cref{sec:model}). For reproducibility, we provide the detailed set of hyperparameters for our best models for each experiment, alongside our code in the corresponding GitHub repository\footnote{https://github.com/lstappen/MuSe2021}, where also a link to the fully trained model weights can be found.
In the following section, we give an overview of all baseline results as summarised in \Cref{tab:base13}. 

\subsection{\musewild}
We evaluated several feature sets and combinations for the prediction of the continuous valence and arousal (\cf \Cref{tab:base13}). The input features \bert in combination with our baseline architecture set to $lr = 0.005$, $h = 128$, and $n = 4$ show superior results for the prediction of valence leading to a CCC of $.4613$ on the development and $.5671$ CCC on test set. For the prediction of arousal, using \ds as input features and setting $lr = 0.001$, $h = 64$, and $n = 2$, yields the best result of all applied systems with a CCC of $.3386$ on the test set. Generally, we found that a unidirectional LSTM-RNN achieves better results for this task than complex bidirectional configurations and is used for the reported \musewild results. When fusing the best performing features of all three modalities \ds, \vggf, and \bert, the late fusion technique reaches $.4863$ and $.5974$ for valence and $.4929$ and $.3257$ for arousal on the development and test set, respectively. This technique yields the highest combined metric (mean of valence and arousal) of $.4616$ (on test) and is our baseline. 

\begin{table*}[t!]
\caption{Reporting Valence, Arousal, Combined ($0.5 \cdot Arousal + 0.5 \cdot Valence$), as well as physical-arousal in CCC for \musewild, \musestress, and \musebio on the devel(opment) and test partitions. For \musesent, we report F1 score across five classes (20\,\% by chance). %Furthermore, the fused psycho-physiological Arousal combining human annotations and EDA is reported in CCC. 
As feature sets, we test \ds, \vgg{}, and \egm{} for audio; \xce, \vggf{} and FAU for video; and \bert for text.
All utilised features are aligned to the label timestamps by imputing missing values or repeating the word embeddings}
% the arousal (Ap) and valence (Vp) predicted signals 

\resizebox{1.0\linewidth}{!}{%linewidth .85
 \begin{tabular}{l|ccc|ccc|ccc|c}
 \toprule
 & \multicolumn{3}{c}{\textbf{\musewild}} & \multicolumn{3}{c}{\textbf{\musesent}} & \multicolumn{3}{c}{\textbf{\musestress}} & \\ 
 Features   & \textbf{Valence}  & \textbf{Arousal}  & \textbf{Combined} & \textbf{Valence}  & \textbf{Arousal}  & \textbf{Combined} & \textbf{Valence}  & \textbf{Arousal}  & \textbf{Combined} & \textbf{\musebio} \\ 
  & devel / test   & devel / test   & devel / test & devel / test   & devel / test   & devel / test & devel / test   & devel / test   & devel / test  & devel / test\\ \hline
 \multicolumn{11}{c}{\textbf{Audio}} \\
 \hline
  \ds   & .1901 / .1019 & .4841 / \textbf{.3386} & .3371 / .2203 & 30.23 / 27.26 & 33.52 / 33.16 & 31.88 / 30.21     & .5018 / .4525 & .3091 / .2341 & .4055 / .3433     & .4423 / .4162 \\
  \vgg  & .1500 / .0054 & .4027 / .2545 & .2764 / .1300         & 30.76 / 25.08 & 36.05 / 31.66 & 33.41 / 28.37     & .5370 / .4766 & .1348 / .0296 & .3359 / .2531       & .3180 / .3967 \\
  \egm  & .1916 / .0019 & .3877 / .2428 & .2897 / .1224         & 32.93 / 25.80 & 36.04 / 31.97 & 34.49 / 28.89     & .5845 / .5018 & .4304 / .4416 & .5075 / .4717       & .3381 / .2416 \\ 
  \hline
 \multicolumn{11}{c}{\textbf{Video}} \\
 \hline
  \xce  & .1872 / .1637 & .2870 / .1793 & .2371 / .1715 & 30.40 / 28.74 & 35.16 / 31.14 & 32.78 / 29.94     & -- / -- & -- / -- & -- / -- & -- / -- \\
  \vggf & .1203 / .1197 & .3201 / .2970 & .2202 / .2084 & 32.29 / 28.86 & 34.57 / 31.32 & 33.43 / 30.09     & .4653 / .4529 & .2004 / .1579 & .3329 / .3054 & .3903 / .4582 \\
  FAU   & .0682 / .1275 & .3045 / .1165 & .1864 / .1220       & 31.37 / 27.38 & 35.21 / 31.43 & 33.29 / 29.41     & .3565 / .2731 & .3313 / .2641 & .3439 / .2686 & .3344 / .1404 \\
  \hline
 \multicolumn{11}{c}{\textbf{Text}} \\
  \hline
 Bert   & .4613 / .5671 & .2716 / .1873 & .3665 / .3772         & 32.68 / 31.90 & \textbf{38.27} / 30.63 & \textbf{35.48} / 31.27     & .2619 / .1747 & .2334 / .1446 & .2477 / .1597 & .2583 / .1604 \\
 \hline
% \multicolumn{10}{c}{\textbf{Early Fusion}} \\
% \hline
%  best A + V & .xxxx / .xxxx      & .xxxx / .xxxx      & .xxxx / .xxxx & .xxxx / .xxxx      & .xxxx / .xxxx      & .xxxx / .xxxx & .6379 / .5785 & .5058 / .4349 & .xxxx / .xxxx & .2298 / .2231 \\
%  best A + T & .xxxx / .xxxx      & .xxxx / .xxxx      & .xxxx / .xxxx & .xxxx / .xxxx      & .xxxx / .xxxx      & .xxxx / .xxxx     & .4400 / .3881 & .3419 / .3904 & .xxxx / .xxxx & .3164 / .3453 \\
%  best V + T & .xxxx / .xxxx      & .xxxx / .xxxx      & .xxxx / .xxxx & .xxxx / .xxxx      & .xxxx / .xxxx      & .xxxx / .xxxx     & .5556 / .4912 & .1760 / .1804 & .xxxx / .xxxx & .2077 / .1897 \\
%  best V + A + T & .xxxx / .xxxx      & .xxxx / .xxxx      & .xxxx / .xxxx & .xxxx / .xxxx      & .xxxx / .xxxx      & .xxxx / .xxxx     & .6061 / .5064 & .2973 / .3463 & .xxxx / .xxxx & .3008 / .2725 \\ 
%  \hline
 \multicolumn{11}{c}{\textbf{Late Fusion}} \\
 \hline
  best A + V     & .2362 / .1220    & .4821 / .2822 & .3592 / .2021            & \textbf{32.96} / 27.92 & 37.72 / \textbf{35.12} & 35.34 / 31.52         & \textbf{.6966} / \textbf{.5614} & \textbf{.5043} / \textbf{.4562} & \textbf{.6005} / \textbf{.5088} & \textbf{.4913 / .4908} \\
  best A + T     & .4782 / .5950    & .4754 / .3046   & .4768 / .4498           & 30.15 / 30.29 & 37.63 / 32.87 & 33.89 / 31.58         & .5684 / .5192 & .4589 / .3227 & .5137 / .4210 & .3931 / .1758 \\
  best V + T     & .4641 / .5874    & .3111 / .1767   & .3876 / .3821           & 30.17 / \textbf{32.91} & 37.51 / 32.73 & 33.84 / \textbf{32.82}         & .5588 / .4250 & .2891 / .1586 & .4240 / .3828 & .2734 / .3000 \\
  best V + A + T & \textbf{.4863 / .5974}    & \textbf{.4929} / .3257   & \textbf{.4896 / .4616}  & 30.37 / 31.01 & 36.72 / 33.20 & 33.55 / 32.11         & .6769 / .5349 & .4819 / .3472 & .5794 / .4411 & .4330 / .3205 \\ 
 
  %Ge + FT + V & .0393 / .0654      & .1809 / .0865     & .1101 / .0760     & .1245 / .1695 \\
 %End2You & FT + VG + RA     & \textbf{.1506 / .2431} & .2587 / .2706 & \textbf{.2047 / .2568} & \textbf{.3198 / .4128} \\ 
 %End2You-Multitask & FT + VG + RA     & -- & -- & -- & .3264 / .4119 \\ \bottomrule 
 % End2You & FT + VG + RA + Ap    & -- & -- & -- & .3318/ .4345 \\ 
 % End2You & FT + VG + RA + Vp    & -- & -- & -- & \textbf{.3450/ .4359}
 \bottomrule 
 \end{tabular}
}
\label{tab:base13}
%\vspace{-0.2cm}
\end{table*}

%\vspace{-0.2cm}
\subsection{\musesent}
\begin{figure}[t!]
    \centering % to be replaced
    \includegraphics[ width=0.45\columnwidth]{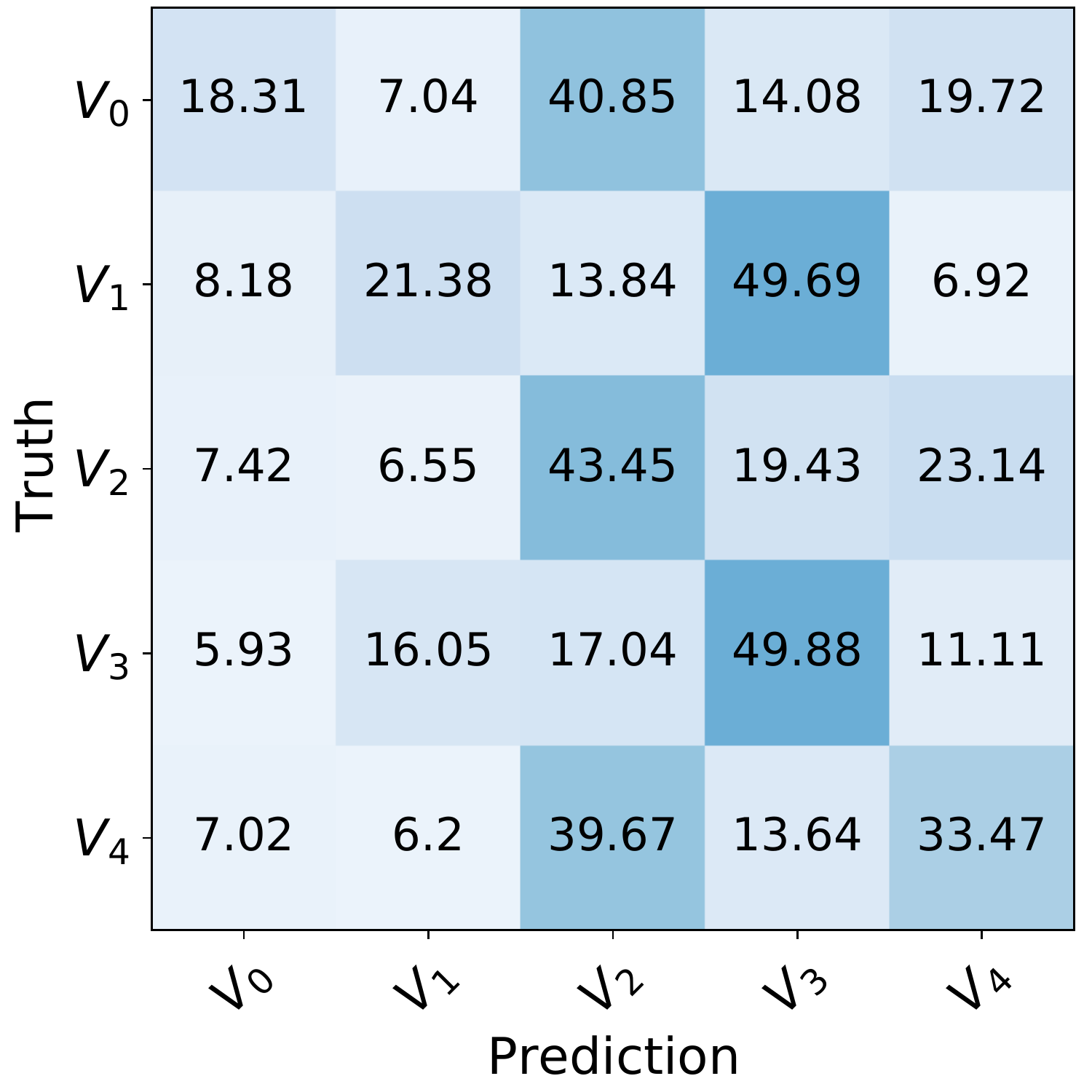}
    \includegraphics[ width=0.45\columnwidth]{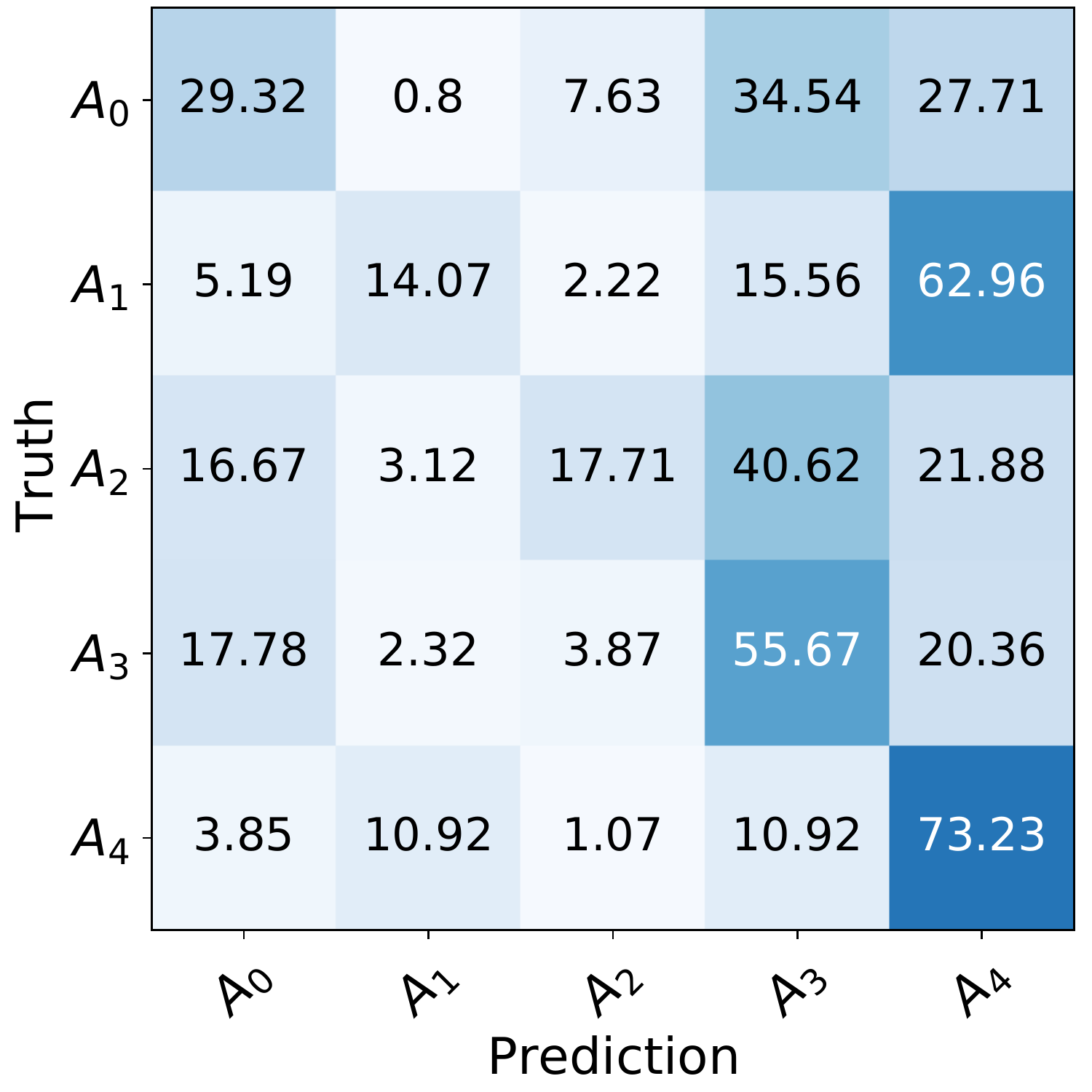}
    %\includegraphics[ width=0.98\columnwidth]{figures/4.pdf} %cm_test_topic.pdf
    % to be replaced by pdf version
    %\includegraphics[width=0.3\linewidth]{figures/arousal_mmt_dsgevg.png} 
    %\includegraphics[width=0.3\linewidth]{figures/XEF_test.png} \\
    %\vspace{-1em}
    \caption{Relative confusion matrices over the 5 valence (left) and arousal (right) classes on the development partition for the \musesent sub-challenge. The results were achieved with the LSTM baseline model using the \egm feature set with hyperparameters of $n$ = 4 (bi-directional), $h$ = 128, and a $lr$ = 0.001 for valence, and for arousal the \bert features with a uni-directional model setting of $n$ = 2, $h$ = 64 and a $lr$ = 0.01.}
    \label{fig:cf}
\end{figure}

For the classification tasks in the \musesent sub-challenge, we give an overview in \Cref{tab:base13} and further provide the confusion matrices for the best uni-modal setups tested on valence and arousal in \Cref{fig:cf}. For the prediction of valence on uni-modal feature inputs, the best result is achieved using the text-based \bert features as input and a baseline model setting of $lr = 0.001$, $h = 64$ and $n = 4$ (bi-directional), with an F1 score of $32.68\,\%$ on the development and $31.90\,\%$ on the test set. Using the audio-based \ds{} features with a $lr = 0.001$, $h = 128$, and $n = 2$ (bi-directional), results in our highest F1 score for arousal with $33.52\,\%$ on the development and $33.16\,\%$ on the test set. Across both targets, we find that LSTM-RNN models with a bidirectional setting and at least two layers tend to achieve better results for this task than smaller architectures. Partially, we see improvements when we apply late fusion. For valence, utilising the predictions of \vggf{} and \bert yields a performance of $32.91\,\%$ F1-score on the test set. For arousal, the audio-visual fusion set-up (\vgg{} and FAU) also improves on the test set, with an F1 score of $35.12\,\%$. Looking at the combined scores (mean of valence and arousal), using the \bert features alone comes out on top for the development set, reaching a $35.48\,\%$ F1 score, while fusing the video- and text-based predictions achieves the highest F1 score of $32.82\,\%$ on the test set.

\subsection{\musestress}
The best results from all feature sets and fusion of modalities are reported in \Cref{tab:base13}\footnote{Of note, besides \egm, we also normalise the \vgg{} features for predicting arousal.}. Having searched the hyperparameter combinations mentioned, we achieve the best results on all settings with a 4-layered unidirectional LSTM equipped with 64-dimensional hidden states and a learning rate of $0.0002$ with a maximum of 100 epochs, and early stopping with a patience of 15 epochs. Here, \egm{} outperforms all other single feature sets for the prediction of valence, achieving $.5845$ CCC on development and $.5018$ CCC on the test set. Regarding arousal, \egm{} is the best scoring single feature set, leading to $.4304$ and $.4416$ CCC on development and test set, respectively. For both valence and arousal prediction, the fusions of the best audio and vision feature sets result in the best performance overall. They achieve CCC values of $.6966$ (development) and $.5614$ (test) for valence, and $.5043$ (development) and $.4562$ (test) for arousal.. It is notable that the text feature set, \bert, performs considerably worse than the best audio and visual features. 
%Consequently, the late fusions of audio and visual features do not benefit from adding the text features. 

We found that, in general, valence reaches a stronger final result than arousal for this task. While this is not surprising for text features, it counters conventional expectations for the audio modality.  A major reason for the poor arousal prediction results may be the TSST scenario, which imitates a job interview. Typically, interviewees try to remain neutral, suppressing nervousness, hence, the arousal shown to their counterpart would be minimal, thus, making arousal more difficult to detect in the \ulm{} data set than other comparable multimodal emotion recognition data sets. 

Although we do not evaluate the provided bio-signal features systematically here, we encourage participants to explore them. To give an example, we achieve $.2495$ and $.1537$ CCC for valence on the development and test sets, respectively, by using only the three provided bio-signals (at a sampling rate of 2\,Hz) as features in a four-layer LSTM. Similarly, they show also promising results for the prediction of arousal, reaching $.1954$ CCC on the development and $.2189$ CCC on test partition. 

%The results for the predictions of arousal as well as valence in \musestress are depicted in \Cref{tab:base13}. 

\subsection{\musebio}
For \musebio, the same LSTM configuration as for \musestress is applied. The results 
%using audio, visual, and textual features as well as fusion variants of them 
are reported in \Cref{tab:base13}. Again, audio and video features considerably outperform the textual BERT features. While BERT only achieves $.2583$ and $.1604$ CCC on development and test data, respectively, the best audio feature set (\ds) leads to $.4423$ and $.4162$ CCC on development and test data, respectively. Consistently, visual features outperform the textual ones, too. The best visual feature set (\vggf) yields $.3903$ and $.4582$ on development and test data respectively and hence shows comparable performance to \ds. Like in \musestress, the late fusion of the best audio (\vgg{}) and video (\vggf{}) predictions yield the best results, namely $.4913$ CCC on development data and $.4908$ CCC on test data. 

Using the one-dimensional biological signals as features might also be beneficial here, even though our model fails to generalise for them. We achieve CCCs of $.4188$  on the development and $.3328$ on the test set using a 4 LSTM layer setting and a learning rate of $0.01$.

%\vspace{-0.7cm}
\section{Conclusions}
In this paper, we introduced MuSe 2021 -- the second Multimodal Sentiment Analysis challenge. MuSe 2021 utilises the \musesec{} multimodal corpus of emotional car reviews and the \ulm{} corpus, including bio-signals, which are newly featured for the MuSe challenge. The 2021 challenge is comprised of four sub-challenges, aimed for predicting in: i) \musewild, the level of the affective dimensions of valence (corresponding to sentiment) and arousal; ii) \musesent, five classes of each, valence and arousal, from video parts containing certain topics; iii) \musestress, the level of continuous valence and arousal from stressful situations; and iv) \musebio a combination of arousal and EDA signals. By intention, we decided to use open-source software to extract a wide range of feature sets to deliver the highest possible transparency and realism for the baselines. Besides the features, we also share the raw data and the developed code for our baselines publicly. The official baseline for each sub-challenge is for \musewild $.5974$ for continuous valence using late fusion and $.3386$ for continuous arousal using \ds{} features; for the five-class classification \musesent, an F1 score of $32.91$\,\% for valence utilising late fusion of vision and text and $35.12$\,\% for arousal utilising a late fusion of audio-video; for \musestress, a CCC of $.5614$ for valence and $.4562$ for arousal, both based on fusion of the best audio and visual features; and finally, for \musebio, a CCC of $.4908$ for physiological-emotion prediction. 

%\textbf{\musewild $.4616$ CCC; \musesent $32.82\,\%$ F1 score; \musestress $.4717$ CCC, and \musebio $.4606$ CCC.}. 
The baselines are improved through the use of a simple fusion method and show the challenge ahead for multimodal sentiment analysis. In the participants' and future efforts, we hope for novel and exciting combinations of the modalities -- such as linking modalities at earlier stages in the pipeline or more closely.
            
%\vspace{-0.25cm}
\section{Acknowledgments}
%\vspace{-0.1cm}
This project has received funding from the European Union's Horizon 2020 research and the DFG's Reinhart Koselleck project No.\ 442218748 (AUDI0NOMOUS). We thank the sponsors of the Challenge, the BMW Group, and audEERING. 

%\td{Organisers, anything to mention here?}

%No.\,115902 (RADAR CNS) and No.\,826506 (sustAGE),  the EPSRC Grant No.\,2021037, and the DFG's Reinhart Koselleckproject No. 442218748 (AUDI0NOMOUS), .


\footnotesize
% \newpage
\bibliographystyle{ACM-Reference-Format}
%\td{check consistency}
\bibliography{sample-base}


\begin{thebibliography}{57}

%%% the Web address.

\ifx \showCODEN    \undefined \def \showCODEN     #1{\unskip}     \fi
\ifx \showDOI      \undefined \def \showDOI       #1{#1}\fi
\ifx \showISBNx    \undefined \def \showISBNx     #1{\unskip}     \fi
\ifx \showISBNxiii \undefined \def \showISBNxiii  #1{\unskip}     \fi
\ifx \showISSN     \undefined \def \showISSN      #1{\unskip}     \fi
\ifx \showLCCN     \undefined \def \showLCCN      #1{\unskip}     \fi
\ifx \shownote     \undefined \def \shownote      #1{#1}          \fi
\ifx \showarticletitle \undefined \def \showarticletitle #1{#1}   \fi
\ifx \showURL      \undefined \def \showURL       {\relax}        \fi
% invisible to TeX
\providecommand\bibfield[2]{#2}
\providecommand\bibinfo[2]{#2}
\providecommand\natexlab[1]{#1}
\providecommand\showeprint[2][]{arXiv:#2}

\bibitem[\protect\citeauthoryear{Amiriparian, Gerczuk, Ottl, Cummins, Freitag,
  Pugachevskiy, Baird, and Schuller}{Amiriparian et~al\mbox{.}}{2017}]%
        {amiriparian2017snore}
\bibfield{author}{\bibinfo{person}{Shahin Amiriparian},
  \bibinfo{person}{Maurice Gerczuk}, \bibinfo{person}{Sandra Ottl},
  \bibinfo{person}{Nicholas Cummins}, \bibinfo{person}{Michael Freitag},
  \bibinfo{person}{Sergey Pugachevskiy}, \bibinfo{person}{Alice Baird}, {and}
  \bibinfo{person}{Bj{\"o}rn~W Schuller}.} \bibinfo{year}{2017}\natexlab{}.
\newblock \showarticletitle{Snore Sound Classification Using Image-Based Deep
  Spectrum Features.}. In \bibinfo{booktitle}{\emph{Proceedings of
  INTERSPEECH}}, Vol.~\bibinfo{volume}{434}. \bibinfo{pages}{3512--3516}.
\newblock


\bibitem[\protect\citeauthoryear{Arevalo, Solorio, Montes-y G{\'o}mez, and
  Gonz{\'a}lez}{Arevalo et~al\mbox{.}}{2020}]%
        {arevalo2020gated}
\bibfield{author}{\bibinfo{person}{John Arevalo}, \bibinfo{person}{Thamar
  Solorio}, \bibinfo{person}{Manuel Montes-y G{\'o}mez}, {and}
  \bibinfo{person}{Fabio~A Gonz{\'a}lez}.} \bibinfo{year}{2020}\natexlab{}.
\newblock \showarticletitle{Gated multimodal networks}.
\newblock \bibinfo{journal}{\emph{Neural Computing and Applications}}
  (\bibinfo{year}{2020}), \bibinfo{pages}{1--20}.
\newblock


\bibitem[\protect\citeauthoryear{Baird, Amiriparian, and Schuller}{Baird
  et~al\mbox{.}}{2019}]%
        {baird2019can}
\bibfield{author}{\bibinfo{person}{Alice Baird}, \bibinfo{person}{Shahin
  Amiriparian}, {and} \bibinfo{person}{Bj{\"o}rn Schuller}.}
  \bibinfo{year}{2019}\natexlab{}.
\newblock \showarticletitle{Can deep generative audio be emotional? Towards an
  approach for personalised emotional audio generation}. In
  \bibinfo{booktitle}{\emph{2019 IEEE 21st International Workshop on Multimedia
  Signal Processing (MMSP)}}. IEEE, \bibinfo{pages}{1--5}.
\newblock


\bibitem[\protect\citeauthoryear{Baird, Stappen, Christ, Schumann, Me{\ss}ner,
  and Schuller}{Baird et~al\mbox{.}}{2021}]%
        {baird2021physiologically}
\bibfield{author}{\bibinfo{person}{Alice Baird}, \bibinfo{person}{Lukas
  Stappen}, \bibinfo{person}{Lukas Christ}, \bibinfo{person}{Lea Schumann},
  \bibinfo{person}{Eva-Maria Me{\ss}ner}, {and} \bibinfo{person}{Bj{\"o}rn~W
  Schuller}.} \bibinfo{year}{2021}\natexlab{}.
\newblock \showarticletitle{A Physiologically-adapted Gold Standard for Arousal
  During a Stress Induced Scenario}. In \bibinfo{booktitle}{\emph{Proceedings
  of the 2nd Multimodal Sentiment Analysis Challenge, co-located with the 29th
  ACM International Conference on Multimedia (ACMMM)}}. ACM,
  \bibinfo{address}{Changu, China}.
\newblock


\bibitem[\protect\citeauthoryear{Baltru\v{s}aitis, Robinson, and
  Morency}{Baltru\v{s}aitis et~al\mbox{.}}{2016}]%
        {Baltruvsaitis16OAO}
\bibfield{author}{\bibinfo{person}{Tadas Baltru\v{s}aitis},
  \bibinfo{person}{Peter Robinson}, {and} \bibinfo{person}{Louis-Philippe
  Morency}.} \bibinfo{year}{2016}\natexlab{}.
\newblock \showarticletitle{{OpenFace: an Open Source Facial Behavior Analysis
  Toolkit}}. In \bibinfo{booktitle}{\emph{Proceedings of the IEEE Winter
  Conference on Applications of Computer Vision}}. \bibinfo{publisher}{IEEE}.
\newblock


\bibitem[\protect\citeauthoryear{Can, Arnrich, and Ersoy}{Can
  et~al\mbox{.}}{2019}]%
        {can2019stress}
\bibfield{author}{\bibinfo{person}{Yekta~Said Can}, \bibinfo{person}{Bert
  Arnrich}, {and} \bibinfo{person}{Cem Ersoy}.}
  \bibinfo{year}{2019}\natexlab{}.
\newblock \showarticletitle{Stress detection in daily life scenarios using
  smart phones and wearable sensors: A survey}.
\newblock \bibinfo{journal}{\emph{Journal of Biomedical Informatics}}
  \bibinfo{volume}{92} (\bibinfo{year}{2019}).
\newblock


\bibitem[\protect\citeauthoryear{Caruelle, Gustafsson, Shams, and
  Lervik-Olsen}{Caruelle et~al\mbox{.}}{2019}]%
        {caruelle2019use}
\bibfield{author}{\bibinfo{person}{Delphine Caruelle}, \bibinfo{person}{Anders
  Gustafsson}, \bibinfo{person}{Poja Shams}, {and} \bibinfo{person}{Line
  Lervik-Olsen}.} \bibinfo{year}{2019}\natexlab{}.
\newblock \showarticletitle{The use of electrodermal activity (EDA) measurement
  to understand consumer emotions--a literature review and a call for action}.
\newblock \bibinfo{journal}{\emph{Journal of Business Research}}
  \bibinfo{volume}{104} (\bibinfo{year}{2019}), \bibinfo{pages}{146--160}.
\newblock


\bibitem[\protect\citeauthoryear{Devlin, Chang, Lee, and Toutanova}{Devlin
  et~al\mbox{.}}{2019}]%
        {devlin2019bert}
\bibfield{author}{\bibinfo{person}{Jacob Devlin}, \bibinfo{person}{Ming-Wei
  Chang}, \bibinfo{person}{Kenton Lee}, {and} \bibinfo{person}{Kristina
  Toutanova}.} \bibinfo{year}{2019}\natexlab{}.
\newblock \showarticletitle{BERT: Pre-training of Deep Bidirectional
  Transformers for Language Understanding}. In
  \bibinfo{booktitle}{\emph{Proceedings of the 2019 Conference of the North
  American Chapter of the Association for Computational Linguistics: Human
  Language Technologies}}. \bibinfo{pages}{4171--4186}.
\newblock


\bibitem[\protect\citeauthoryear{Dhall, Sharma, Goecke, and Gedeon}{Dhall
  et~al\mbox{.}}{2020}]%
        {dhall2020emotiw}
\bibfield{author}{\bibinfo{person}{Abhinav Dhall}, \bibinfo{person}{Garima
  Sharma}, \bibinfo{person}{Roland Goecke}, {and} \bibinfo{person}{Tom
  Gedeon}.} \bibinfo{year}{2020}\natexlab{}.
\newblock \showarticletitle{Emotiw 2020: Driver gaze, group emotion, student
  engagement and physiological signal based challenges}. In
  \bibinfo{booktitle}{\emph{Proceedings of the 2020 International Conference on
  Multimodal Interaction (ICMI)}}. \bibinfo{pages}{784--789}.
\newblock


\bibitem[\protect\citeauthoryear{Duda, Hart, et~al\mbox{.}}{Duda
  et~al\mbox{.}}{1973}]%
        {duda1973pattern}
\bibfield{author}{\bibinfo{person}{Richard~O Duda}, \bibinfo{person}{Peter~E
  Hart}, {et~al\mbox{.}}} \bibinfo{year}{1973}\natexlab{}.
\newblock \bibinfo{booktitle}{\emph{Pattern classification and scene
  analysis}}. Vol.~\bibinfo{volume}{3}.
\newblock \bibinfo{publisher}{Wiley New York}.
\newblock


\bibitem[\protect\citeauthoryear{{Egede}, {Song}, {Olugbade}, {Wang},
  {Williams}, {Meng}, {Aung}, {Lane}, {Valstar}, and
  {Bianchi-Berthouze}}{{Egede} et~al\mbox{.}}{2020}]%
        {egede2020FG}
\bibfield{author}{\bibinfo{person}{J.~O. {Egede}}, \bibinfo{person}{S. {Song}},
  \bibinfo{person}{T.~A. {Olugbade}}, \bibinfo{person}{C. {Wang}},
  \bibinfo{person}{A.~C. D.~C. {Williams}}, \bibinfo{person}{H. {Meng}},
  \bibinfo{person}{M. {Aung}}, \bibinfo{person}{N.~D. {Lane}},
  \bibinfo{person}{M. {Valstar}}, {and} \bibinfo{person}{N.
  {Bianchi-Berthouze}}.} \bibinfo{year}{2020}\natexlab{}.
\newblock \showarticletitle{EMOPAIN Challenge 2020: Multimodal Pain Evaluation
  from Facial and Bodily Expressions}. In \bibinfo{booktitle}{\emph{2020 15th
  IEEE International Conference on Automatic Face and Gesture Recognition (FG
  2020)}}. \bibinfo{pages}{849--856}.
\newblock


\bibitem[\protect\citeauthoryear{Eyben, Scherer, Schuller, Sundberg, Andr{\'e},
  Busso, Devillers, Epps, Laukka, Narayanan, et~al\mbox{.}}{Eyben
  et~al\mbox{.}}{2015}]%
        {eyben2015geneva}
\bibfield{author}{\bibinfo{person}{Florian Eyben}, \bibinfo{person}{Klaus~R
  Scherer}, \bibinfo{person}{Bj{\"o}rn~W Schuller}, \bibinfo{person}{Johan
  Sundberg}, \bibinfo{person}{Elisabeth Andr{\'e}}, \bibinfo{person}{Carlos
  Busso}, \bibinfo{person}{Laurence~Y Devillers}, \bibinfo{person}{Julien
  Epps}, \bibinfo{person}{Petri Laukka}, \bibinfo{person}{Shrikanth~S
  Narayanan}, {et~al\mbox{.}}} \bibinfo{year}{2015}\natexlab{}.
\newblock \showarticletitle{The Geneva minimalistic acoustic parameter set
  (GeMAPS) for voice research and affective computing}.
\newblock \bibinfo{journal}{\emph{IEEE Transactions on Affective Computing}}
  \bibinfo{volume}{7}, \bibinfo{number}{2} (\bibinfo{year}{2015}),
  \bibinfo{pages}{190--202}.
\newblock


\bibitem[\protect\citeauthoryear{Eyben, W{\"o}llmer, and Schuller}{Eyben
  et~al\mbox{.}}{2010}]%
        {eyben2010opensmile}
\bibfield{author}{\bibinfo{person}{Florian Eyben}, \bibinfo{person}{Martin
  W{\"o}llmer}, {and} \bibinfo{person}{Bj{\"o}rn Schuller}.}
  \bibinfo{year}{2010}\natexlab{}.
\newblock \showarticletitle{Opensmile: the munich versatile and fast
  open-source audio feature extractor}. In
  \bibinfo{booktitle}{\emph{Proceedings of the 18th ACM International
  Conference on Multimedia}}. \bibinfo{pages}{1459--1462}.
\newblock


\bibitem[\protect\citeauthoryear{Gemmeke, Ellis, Freedman, Jansen, Lawrence,
  Moore, Plakal, and Ritter}{Gemmeke et~al\mbox{.}}{2017}]%
        {gemmeke2017audio}
\bibfield{author}{\bibinfo{person}{Jort~F Gemmeke}, \bibinfo{person}{Daniel~PW
  Ellis}, \bibinfo{person}{Dylan Freedman}, \bibinfo{person}{Aren Jansen},
  \bibinfo{person}{Wade Lawrence}, \bibinfo{person}{R~Channing Moore},
  \bibinfo{person}{Manoj Plakal}, {and} \bibinfo{person}{Marvin Ritter}.}
  \bibinfo{year}{2017}\natexlab{}.
\newblock \showarticletitle{Audio set: An ontology and human-labeled dataset
  for audio events}. In \bibinfo{booktitle}{\emph{2017 IEEE International
  Conference on Acoustics, Speech and Signal Processing (ICASSP)}}. IEEE,
  \bibinfo{pages}{776--780}.
\newblock


\bibitem[\protect\citeauthoryear{Gomez, Gibert, Gomez, and Karatzas}{Gomez
  et~al\mbox{.}}{2020}]%
        {gomez2020exploring}
\bibfield{author}{\bibinfo{person}{Raul Gomez}, \bibinfo{person}{Jaume Gibert},
  \bibinfo{person}{Lluis Gomez}, {and} \bibinfo{person}{Dimosthenis Karatzas}.}
  \bibinfo{year}{2020}\natexlab{}.
\newblock \showarticletitle{Exploring Hate Speech Detection in Multimodal
  Publications}. In \bibinfo{booktitle}{\emph{The IEEE Winter Conference on
  Applications of Computer Vision}}. \bibinfo{pages}{1470--1478}.
\newblock


\bibitem[\protect\citeauthoryear{Grimm and Kroschel}{Grimm and
  Kroschel}{2005}]%
        {grimm2005evaluation}
\bibfield{author}{\bibinfo{person}{Michael Grimm} {and}
  \bibinfo{person}{Kristian Kroschel}.} \bibinfo{year}{2005}\natexlab{}.
\newblock \showarticletitle{Evaluation of natural emotions using self
  assessment manikins}. In \bibinfo{booktitle}{\emph{IEEE Workshop on Automatic
  Speech Recognition and Understanding, 2005.}} IEEE,
  \bibinfo{pages}{381--385}.
\newblock


\bibitem[\protect\citeauthoryear{He, Zhang, Ren, and Sun}{He
  et~al\mbox{.}}{2016}]%
        {he2016deep}
\bibfield{author}{\bibinfo{person}{Kaiming He}, \bibinfo{person}{Xiangyu
  Zhang}, \bibinfo{person}{Shaoqing Ren}, {and} \bibinfo{person}{Jian Sun}.}
  \bibinfo{year}{2016}\natexlab{}.
\newblock \showarticletitle{Deep residual learning for image recognition}. In
  \bibinfo{booktitle}{\emph{Proceedings of the IEEE Conference on Computer
  Vision and Pattern Recognition}}. \bibinfo{pages}{770--778}.
\newblock


\bibitem[\protect\citeauthoryear{Hershey, Chaudhuri, Ellis, Gemmeke, Jansen,
  Moore, Plakal, Platt, Saurous, Seybold, et~al\mbox{.}}{Hershey
  et~al\mbox{.}}{2017}]%
        {hershey2017cnn}
\bibfield{author}{\bibinfo{person}{Shawn Hershey}, \bibinfo{person}{Sourish
  Chaudhuri}, \bibinfo{person}{Daniel~PW Ellis}, \bibinfo{person}{Jort~F
  Gemmeke}, \bibinfo{person}{Aren Jansen}, \bibinfo{person}{R~Channing Moore},
  \bibinfo{person}{Manoj Plakal}, \bibinfo{person}{Devin Platt},
  \bibinfo{person}{Rif~A Saurous}, \bibinfo{person}{Bryan Seybold},
  {et~al\mbox{.}}} \bibinfo{year}{2017}\natexlab{}.
\newblock \showarticletitle{CNN architectures for large-scale audio
  classification}. In \bibinfo{booktitle}{\emph{2017 IEEE International
  Conference on Acoustics, Speech and Signal Processing (ICASSP)}}. IEEE,
  \bibinfo{pages}{131--135}.
\newblock


\bibitem[\protect\citeauthoryear{Hoffmann, Scheck, Schuster, Walter, Limbrecht,
  Traue, and Kessler}{Hoffmann et~al\mbox{.}}{2012}]%
        {hoffmann2012mapping}
\bibfield{author}{\bibinfo{person}{Holger Hoffmann}, \bibinfo{person}{Andreas
  Scheck}, \bibinfo{person}{Timo Schuster}, \bibinfo{person}{Steffen Walter},
  \bibinfo{person}{Kerstin Limbrecht}, \bibinfo{person}{Harald~C Traue}, {and}
  \bibinfo{person}{Henrik Kessler}.} \bibinfo{year}{2012}\natexlab{}.
\newblock \showarticletitle{Mapping discrete emotions into the dimensional
  space: An empirical approach}. In \bibinfo{booktitle}{\emph{2012 IEEE
  International Conference on Systems, Man, and Cybernetics (SMC)}}. IEEE,
  \bibinfo{pages}{3316--3320}.
\newblock


\bibitem[\protect\citeauthoryear{Kirschbaum, Pirke, and Hellhammer}{Kirschbaum
  et~al\mbox{.}}{1993}]%
        {kirschbaum1993trier}
\bibfield{author}{\bibinfo{person}{Clemens Kirschbaum},
  \bibinfo{person}{Karl-Martin Pirke}, {and} \bibinfo{person}{Dirk~H
  Hellhammer}.} \bibinfo{year}{1993}\natexlab{}.
\newblock \showarticletitle{The ‘Trier Social Stress Test’--a tool for
  investigating psychobiological stress responses in a laboratory setting}.
\newblock \bibinfo{journal}{\emph{Neuropsychobiology}} \bibinfo{volume}{28},
  \bibinfo{number}{1-2} (\bibinfo{year}{1993}), \bibinfo{pages}{76--81}.
\newblock


\bibitem[\protect\citeauthoryear{Kollias, Schulc, Hajiyev, and
  Zafeiriou}{Kollias et~al\mbox{.}}{2020}]%
        {kollias2020analysing}
\bibfield{author}{\bibinfo{person}{Dimitrios Kollias}, \bibinfo{person}{Attila
  Schulc}, \bibinfo{person}{Elnar Hajiyev}, {and} \bibinfo{person}{Stefanos
  Zafeiriou}.} \bibinfo{year}{2020}\natexlab{}.
\newblock \showarticletitle{Analysing affective behavior in the first ABAW 2020
  competition}.
\newblock \bibinfo{journal}{\emph{arXiv preprint arXiv:2001.11409}}
  (\bibinfo{year}{2020}).
\newblock


\bibitem[\protect\citeauthoryear{Liu, Li, Xiong, Gao, and Wu}{Liu
  et~al\mbox{.}}{2010}]%
        {liu2010understanding}
\bibfield{author}{\bibinfo{person}{Yanchi Liu}, \bibinfo{person}{Zhongmou Li},
  \bibinfo{person}{Hui Xiong}, \bibinfo{person}{Xuedong Gao}, {and}
  \bibinfo{person}{Junjie Wu}.} \bibinfo{year}{2010}\natexlab{}.
\newblock \showarticletitle{Understanding of internal clustering validation
  measures}. In \bibinfo{booktitle}{\emph{Proceedings of the IEEE International
  Conference on Data Mining}}. IEEE, \bibinfo{pages}{911--916}.
\newblock


\bibitem[\protect\citeauthoryear{Liu, Luo, Wang, and Tang}{Liu
  et~al\mbox{.}}{2015}]%
        {liu2015faceattributes}
\bibfield{author}{\bibinfo{person}{Ziwei Liu}, \bibinfo{person}{Ping Luo},
  \bibinfo{person}{Xiaogang Wang}, {and} \bibinfo{person}{Xiaoou Tang}.}
  \bibinfo{year}{2015}\natexlab{}.
\newblock \showarticletitle{Deep Learning Face Attributes in the Wild}. In
  \bibinfo{booktitle}{\emph{Proceedings of International Conference on Computer
  Vision (ICCV)}}.
\newblock


\bibitem[\protect\citeauthoryear{Lloyd}{Lloyd}{1982}]%
        {lloyd1982least}
\bibfield{author}{\bibinfo{person}{Stuart Lloyd}.}
  \bibinfo{year}{1982}\natexlab{}.
\newblock \showarticletitle{Least squares quantization in PCM}.
\newblock \bibinfo{journal}{\emph{IEEE Transactions on Information Theory}}
  \bibinfo{volume}{28}, \bibinfo{number}{2} (\bibinfo{year}{1982}),
  \bibinfo{pages}{129--137}.
\newblock


\bibitem[\protect\citeauthoryear{McAuliffe, Socolof, Mihuc, Wagner, and
  Sonderegger}{McAuliffe et~al\mbox{.}}{2017}]%
        {mcauliffe2017montreal}
\bibfield{author}{\bibinfo{person}{Michael McAuliffe},
  \bibinfo{person}{Michaela Socolof}, \bibinfo{person}{Sarah Mihuc},
  \bibinfo{person}{Michael Wagner}, {and} \bibinfo{person}{Morgan
  Sonderegger}.} \bibinfo{year}{2017}\natexlab{}.
\newblock \showarticletitle{Montreal Forced Aligner: Trainable Text-Speech
  Alignment Using Kaldi.}. In \bibinfo{booktitle}{\emph{Proceedings of
  INTERSPEECH}}, Vol.~\bibinfo{volume}{2017}. \bibinfo{pages}{498--502}.
\newblock


\bibitem[\protect\citeauthoryear{Mohammad}{Mohammad}{2016}]%
        {mohammad2016sentiment}
\bibfield{author}{\bibinfo{person}{Saif~M Mohammad}.}
  \bibinfo{year}{2016}\natexlab{}.
\newblock \showarticletitle{Sentiment analysis: Detecting valence, emotions,
  and other affectual states from text}.
\newblock In \bibinfo{booktitle}{\emph{Emotion Measurement}}.
  \bibinfo{publisher}{Elsevier}, \bibinfo{pages}{201--237}.
\newblock


\bibitem[\protect\citeauthoryear{Nicolaou, Pavlovic, and Pantic}{Nicolaou
  et~al\mbox{.}}{2014}]%
        {nicolaou2014dynamic}
\bibfield{author}{\bibinfo{person}{Mihalis~A Nicolaou},
  \bibinfo{person}{Vladimir Pavlovic}, {and} \bibinfo{person}{Maja Pantic}.}
  \bibinfo{year}{2014}\natexlab{}.
\newblock \showarticletitle{Dynamic probabilistic cca for analysis of affective
  behavior and fusion of continuous annotations}.
\newblock \bibinfo{journal}{\emph{IEEE Transactions on Pattern Analysis and
  Machine Intelligence}} \bibinfo{volume}{36}, \bibinfo{number}{7}
  (\bibinfo{year}{2014}), \bibinfo{pages}{1299--1311}.
\newblock


\bibitem[\protect\citeauthoryear{Panagakis, Nicolaou, Zafeiriou, and
  Pantic}{Panagakis et~al\mbox{.}}{2015}]%
        {panagakis2015robust}
\bibfield{author}{\bibinfo{person}{Yannis Panagakis},
  \bibinfo{person}{Mihalis~A Nicolaou}, \bibinfo{person}{Stefanos Zafeiriou},
  {and} \bibinfo{person}{Maja Pantic}.} \bibinfo{year}{2015}\natexlab{}.
\newblock \showarticletitle{Robust correlated and individual component
  analysis}.
\newblock \bibinfo{journal}{\emph{IEEE Transactions on Pattern Analysis and
  Machine Intelligence}} \bibinfo{volume}{38}, \bibinfo{number}{8}
  (\bibinfo{year}{2015}), \bibinfo{pages}{1665--1678}.
\newblock


\bibitem[\protect\citeauthoryear{Pandit and Schuller}{Pandit and
  Schuller}{2019}]%
        {pandit2019many}
\bibfield{author}{\bibinfo{person}{Vedhas Pandit} {and}
  \bibinfo{person}{Bj{\"o}rn Schuller}.} \bibinfo{year}{2019}\natexlab{}.
\newblock \showarticletitle{On Many-to-Many Mapping Between Concordance
  Correlation Coefficient and Mean Square Error}.
\newblock \bibinfo{journal}{\emph{arXiv preprint arXiv:1902.05180}}
  (\bibinfo{year}{2019}).
\newblock


\bibitem[\protect\citeauthoryear{Parkhi, Vedaldi, and Zisserman}{Parkhi
  et~al\mbox{.}}{2015}]%
        {Omkar2015recognition}
\bibfield{author}{\bibinfo{person}{Omkar~M. Parkhi}, \bibinfo{person}{Andrea
  Vedaldi}, {and} \bibinfo{person}{Andrew Zisserman}.}
  \bibinfo{year}{2015}\natexlab{}.
\newblock \showarticletitle{Deep Face Recognition}. In
  \bibinfo{booktitle}{\emph{Proceedings of the British Machine Vision
  Conference (BMVC)}}. \bibinfo{pages}{41.1--41.12}.
\newblock


\bibitem[\protect\citeauthoryear{Pourmohammadi and Maleki}{Pourmohammadi and
  Maleki}{2020}]%
        {pourmohammadi2020stress}
\bibfield{author}{\bibinfo{person}{Sara Pourmohammadi} {and}
  \bibinfo{person}{Ali Maleki}.} \bibinfo{year}{2020}\natexlab{}.
\newblock \showarticletitle{Stress detection using ECG and EMG signals: A
  comprehensive study}.
\newblock \bibinfo{journal}{\emph{Computer Methods and Programs in
  Biomedicine}}  \bibinfo{volume}{193} (\bibinfo{year}{2020}),
  \bibinfo{pages}{105482}.
\newblock


\bibitem[\protect\citeauthoryear{Preo{\c{t}}iuc-Pietro, Schwartz, Park,
  Eichstaedt, Kern, Ungar, and Shulman}{Preo{\c{t}}iuc-Pietro
  et~al\mbox{.}}{2016}]%
        {preoctiuc2016modelling}
\bibfield{author}{\bibinfo{person}{Daniel Preo{\c{t}}iuc-Pietro},
  \bibinfo{person}{H~Andrew Schwartz}, \bibinfo{person}{Gregory Park},
  \bibinfo{person}{Johannes Eichstaedt}, \bibinfo{person}{Margaret Kern},
  \bibinfo{person}{Lyle Ungar}, {and} \bibinfo{person}{Elisabeth Shulman}.}
  \bibinfo{year}{2016}\natexlab{}.
\newblock \showarticletitle{Modelling valence and arousal in facebook posts}.
  In \bibinfo{booktitle}{\emph{Proceedings of the 7th Workshop on Computational
  Approaches to Subjectivity, Sentiment and Social Media Analysis}}.
  \bibinfo{pages}{9--15}.
\newblock


\bibitem[\protect\citeauthoryear{Qiu, Feng, Yang, and Tian}{Qiu
  et~al\mbox{.}}{2020}]%
        {qiu2020multimodal}
\bibfield{author}{\bibinfo{person}{Xiaoyu Qiu}, \bibinfo{person}{Zhiquan Feng},
  \bibinfo{person}{Xiaohui Yang}, {and} \bibinfo{person}{Jinglan Tian}.}
  \bibinfo{year}{2020}\natexlab{}.
\newblock \showarticletitle{Multimodal Fusion of Speech and Gesture Recognition
  based on Deep Learning}. In \bibinfo{booktitle}{\emph{Journal of Physics:
  Conference Series}}, Vol.~\bibinfo{volume}{1453}.
\newblock


\bibitem[\protect\citeauthoryear{Ringeval, Schuller, Valstar, Cowie, Kaya,
  Schmitt, Amiriparian, Cummins, Lalanne, Michaud, et~al\mbox{.}}{Ringeval
  et~al\mbox{.}}{2018}]%
        {ringeval2018avec}
\bibfield{author}{\bibinfo{person}{Fabien Ringeval}, \bibinfo{person}{Bj{\"o}rn
  Schuller}, \bibinfo{person}{Michel Valstar}, \bibinfo{person}{Roddy Cowie},
  \bibinfo{person}{Heysem Kaya}, \bibinfo{person}{Maximilian Schmitt},
  \bibinfo{person}{Shahin Amiriparian}, \bibinfo{person}{Nicholas Cummins},
  \bibinfo{person}{Denis Lalanne}, \bibinfo{person}{Adrien Michaud},
  {et~al\mbox{.}}} \bibinfo{year}{2018}\natexlab{}.
\newblock \showarticletitle{AVEC 2018 workshop and challenge: Bipolar disorder
  and cross-cultural affect recognition}. In
  \bibinfo{booktitle}{\emph{Proceedings of the 2018 on Audio/Visual Emotion
  Challenge and Workshop}}. \bibinfo{pages}{3--13}.
\newblock


\bibitem[\protect\citeauthoryear{Ringeval, Schuller, Valstar, Gratch, Cowie,
  Scherer, Mozgai, Cummins, Schmitt, and Pantic}{Ringeval
  et~al\mbox{.}}{2017}]%
        {ringeval2017avec}
\bibfield{author}{\bibinfo{person}{Fabien Ringeval}, \bibinfo{person}{Bj{\"o}rn
  Schuller}, \bibinfo{person}{Michel Valstar}, \bibinfo{person}{Jonathan
  Gratch}, \bibinfo{person}{Roddy Cowie}, \bibinfo{person}{Stefan Scherer},
  \bibinfo{person}{Sharon Mozgai}, \bibinfo{person}{Nicholas Cummins},
  \bibinfo{person}{Maximilian Schmitt}, {and} \bibinfo{person}{Maja Pantic}.}
  \bibinfo{year}{2017}\natexlab{}.
\newblock \showarticletitle{Avec 2017: Real-life depression, and affect
  recognition workshop and challenge}. In \bibinfo{booktitle}{\emph{Proceedings
  of the 7th Annual Workshop on Audio/Visual Emotion Challenge}}.
  \bibinfo{pages}{3--9}.
\newblock


\bibitem[\protect\citeauthoryear{Rousseeuw}{Rousseeuw}{1987}]%
        {rousseeuw1987silhouettes}
\bibfield{author}{\bibinfo{person}{Peter~J Rousseeuw}.}
  \bibinfo{year}{1987}\natexlab{}.
\newblock \showarticletitle{Silhouettes: a graphical aid to the interpretation
  and validation of cluster analysis}.
\newblock \bibinfo{journal}{\emph{J. Comput. Appl. Math.}}
  \bibinfo{volume}{20} (\bibinfo{year}{1987}), \bibinfo{pages}{53--65}.
\newblock


\bibitem[\protect\citeauthoryear{Schuller, Batliner, Bergler, Mascolo, Han,
  Lefter, Kaya, Amiriparian, Baird, Stappen, et~al\mbox{.}}{Schuller
  et~al\mbox{.}}{2021}]%
        {schuller2021interspeech}
\bibfield{author}{\bibinfo{person}{Bj{\"o}rn~W Schuller},
  \bibinfo{person}{Anton Batliner}, \bibinfo{person}{Christian Bergler},
  \bibinfo{person}{Cecilia Mascolo}, \bibinfo{person}{Jing Han},
  \bibinfo{person}{Iulia Lefter}, \bibinfo{person}{Heysem Kaya},
  \bibinfo{person}{Shahin Amiriparian}, \bibinfo{person}{Alice Baird},
  \bibinfo{person}{Lukas Stappen}, {et~al\mbox{.}}}
  \bibinfo{year}{2021}\natexlab{}.
\newblock \showarticletitle{The INTERSPEECH 2021 Computational Paralinguistics
  Challenge: COVID-19 cough, COVID-19 speech, escalation \& primates}.
\newblock \bibinfo{journal}{\emph{arXiv preprint arXiv:2102.13468}}
  (\bibinfo{year}{2021}).
\newblock


\bibitem[\protect\citeauthoryear{Schuller, Batliner, Bergler, Messner,
  Hamilton, Amiriparian, Baird, Rizos, Schmitt, Stappen,
  et~al\mbox{.}}{Schuller et~al\mbox{.}}{2020}]%
        {schuller2020interspeech}
\bibfield{author}{\bibinfo{person}{Bj{\"o}rn~W Schuller},
  \bibinfo{person}{Anton Batliner}, \bibinfo{person}{Christian Bergler},
  \bibinfo{person}{Eva-Maria Messner}, \bibinfo{person}{Antonia Hamilton},
  \bibinfo{person}{Shahin Amiriparian}, \bibinfo{person}{Alice Baird},
  \bibinfo{person}{Georgios Rizos}, \bibinfo{person}{Maximilian Schmitt},
  \bibinfo{person}{Lukas Stappen}, {et~al\mbox{.}}}
  \bibinfo{year}{2020}\natexlab{}.
\newblock \showarticletitle{The INTERSPEECH 2020 Computational Paralinguistics
  Challenge: Elderly Emotion, Breathing \& Masks}.
\newblock \bibinfo{journal}{\emph{Proceedings of INTERSPEECH}}
  (\bibinfo{year}{2020}).
\newblock


\bibitem[\protect\citeauthoryear{Schuller, Steidl, Batliner, Marschik,
  Baumeister, Dong, Hantke, Pokorny, Rathner, Bartl-Pokorny,
  et~al\mbox{.}}{Schuller et~al\mbox{.}}{2018}]%
        {schuller2018interspeech}
\bibfield{author}{\bibinfo{person}{Bj{\"o}rn~W Schuller},
  \bibinfo{person}{Stefan Steidl}, \bibinfo{person}{Anton Batliner},
  \bibinfo{person}{Peter~B Marschik}, \bibinfo{person}{Harald Baumeister},
  \bibinfo{person}{Fengquan Dong}, \bibinfo{person}{Simone Hantke},
  \bibinfo{person}{Florian~B Pokorny}, \bibinfo{person}{Eva-Maria Rathner},
  \bibinfo{person}{Katrin~D Bartl-Pokorny}, {et~al\mbox{.}}}
  \bibinfo{year}{2018}\natexlab{}.
\newblock \showarticletitle{The INTERSPEECH 2018 Computational Paralinguistics
  Challenge: Atypical \& Self-Assessed Affect, Crying \& Heart Beats.}. In
  \bibinfo{booktitle}{\emph{Proceedings of INTERSPEECH}}.
  \bibinfo{pages}{122--126}.
\newblock


\bibitem[\protect\citeauthoryear{Shukla, Barreda-Angeles, Oliver, Nandi, and
  Puig}{Shukla et~al\mbox{.}}{2019}]%
        {shukla2019feature}
\bibfield{author}{\bibinfo{person}{Jainendra Shukla}, \bibinfo{person}{Miguel
  Barreda-Angeles}, \bibinfo{person}{Joan Oliver}, \bibinfo{person}{GC Nandi},
  {and} \bibinfo{person}{Domenec Puig}.} \bibinfo{year}{2019}\natexlab{}.
\newblock \showarticletitle{Feature extraction and selection for emotion
  recognition from electrodermal activity}.
\newblock \bibinfo{journal}{\emph{IEEE Transactions on Affective Computing}}
  (\bibinfo{year}{2019}).
\newblock


\bibitem[\protect\citeauthoryear{Simonyan and Zisserman}{Simonyan and
  Zisserman}{2014}]%
        {simonyan2014very}
\bibfield{author}{\bibinfo{person}{Karen Simonyan} {and}
  \bibinfo{person}{Andrew Zisserman}.} \bibinfo{year}{2014}\natexlab{}.
\newblock \showarticletitle{Very deep convolutional networks for large-scale
  image recognition}.
\newblock \bibinfo{journal}{\emph{arXiv preprint arXiv:1409.1556}}
  (\bibinfo{year}{2014}).
\newblock


\bibitem[\protect\citeauthoryear{Stappen, Baird, Rizos, Tzirakis, Du, Hafner,
  Schumann, Mallol-Ragolta, Schuller, Lefter, Cambria, and
  Kompatsiaris}{Stappen et~al\mbox{.}}{2020a}]%
        {stappen2020muse1}
\bibfield{author}{\bibinfo{person}{Lukas Stappen}, \bibinfo{person}{Alice
  Baird}, \bibinfo{person}{Georgios Rizos}, \bibinfo{person}{Panagiotis
  Tzirakis}, \bibinfo{person}{Xinchen Du}, \bibinfo{person}{Felix Hafner},
  \bibinfo{person}{Lea Schumann}, \bibinfo{person}{Adria Mallol-Ragolta},
  \bibinfo{person}{Bjoern~W. Schuller}, \bibinfo{person}{Iulia Lefter},
  \bibinfo{person}{Erik Cambria}, {and} \bibinfo{person}{Ioannis
  Kompatsiaris}.} \bibinfo{year}{2020}\natexlab{a}.
\newblock \showarticletitle{MuSe 2020 Challenge and Workshop: Multimodal
  Sentiment Analysis, Emotion-Target Engagement and Trustworthiness Detection
  in Real-Life Media}. In \bibinfo{booktitle}{\emph{Proceedings of the 1st
  International on Multimodal Sentiment Analysis in Real-Life Media Challenge
  and Workshop}}. ACM, \bibinfo{pages}{35–44}.
\newblock


\bibitem[\protect\citeauthoryear{Stappen, Baird, Schumann, and
  Schuller}{Stappen et~al\mbox{.}}{2021a}]%
        {stappen2021multimodal}
\bibfield{author}{\bibinfo{person}{Lukas Stappen}, \bibinfo{person}{Alice
  Baird}, \bibinfo{person}{Lea Schumann}, {and} \bibinfo{person}{Björn
  Schuller}.} \bibinfo{year}{2021}\natexlab{a}.
\newblock \showarticletitle{The Multimodal Sentiment Analysis in Car Reviews
  (MuSe-CaR) Dataset: Collection, Insights and Improvements}.
\newblock \bibinfo{journal}{\emph{IEEE Transactions on Affective Computing
  (Early Access)}} (\bibinfo{date}{June} \bibinfo{year}{2021}).
\newblock
\showISSN{1949-3045}
\urldef\tempurl%
\url{https://doi.org/10.1109/TAFFC.2021.3097002}
\showDOI{\tempurl}


\bibitem[\protect\citeauthoryear{Stappen, Brunn, and Schuller}{Stappen
  et~al\mbox{.}}{2020b}]%
        {stappen2020cross}
\bibfield{author}{\bibinfo{person}{Lukas Stappen}, \bibinfo{person}{Fabian
  Brunn}, {and} \bibinfo{person}{Bj{\"o}rn Schuller}.}
  \bibinfo{year}{2020}\natexlab{b}.
\newblock \showarticletitle{Cross-lingual zero-and few-shot hate speech
  detection utilising frozen transformer language models and AXEL}.
\newblock \bibinfo{journal}{\emph{arXiv preprint arXiv:2004.13850}}
  (\bibinfo{year}{2020}).
\newblock


\bibitem[\protect\citeauthoryear{Stappen, Karas, Cummins, Ringeval, Scherer,
  and Schuller}{Stappen et~al\mbox{.}}{2019}]%
        {stappen2019speech}
\bibfield{author}{\bibinfo{person}{Lukas Stappen}, \bibinfo{person}{Vincent
  Karas}, \bibinfo{person}{Nicholas Cummins}, \bibinfo{person}{Fabien
  Ringeval}, \bibinfo{person}{Klaus Scherer}, {and} \bibinfo{person}{Bj{\"o}rn
  Schuller}.} \bibinfo{year}{2019}\natexlab{}.
\newblock \showarticletitle{From speech to facial activity: towards cross-modal
  sequence-to-sequence attention networks}. In \bibinfo{booktitle}{\emph{2019
  IEEE 21st International Workshop on Multimedia Signal Processing (MMSP)}}.
  IEEE, \bibinfo{pages}{1--6}.
\newblock


\bibitem[\protect\citeauthoryear{Stappen, Meßner, Cambria, Zhao, and
  Schuller}{Stappen et~al\mbox{.}}{2021b}]%
        {stappen2021summary}
\bibfield{author}{\bibinfo{person}{Lukas Stappen}, \bibinfo{person}{Eva-Maria
  Meßner}, \bibinfo{person}{Erik Cambria}, \bibinfo{person}{Guoying Zhao},
  {and} \bibinfo{person}{Björn~W. Schuller}.}
  \bibinfo{year}{2021}\natexlab{b}.
\newblock \showarticletitle{MuSe 2021 Challenge: Multimodal Emotion,
  Sentiment,Physiological-Emotion, and Stress Detection}. In
  \bibinfo{booktitle}{\emph{29th ACM International Conference on Multimedia
  (ACMMM)}}. ACM, \bibinfo{address}{Virtual Event, China}.
\newblock


\bibitem[\protect\citeauthoryear{Stappen, Rizos, Hasan, Hain, and
  Schuller}{Stappen et~al\mbox{.}}{2020c}]%
        {stappen2020uncertainty}
\bibfield{author}{\bibinfo{person}{Lukas Stappen}, \bibinfo{person}{Georgios
  Rizos}, \bibinfo{person}{Madina Hasan}, \bibinfo{person}{Thomas Hain}, {and}
  \bibinfo{person}{Bj{\"o}rn~W Schuller}.} \bibinfo{year}{2020}\natexlab{c}.
\newblock \showarticletitle{Uncertainty-Aware Machine Support for Paper
  Reviewing on the INTERSPEECH 2019 Submission Corpus}.
\newblock \bibinfo{journal}{\emph{Proceedings of INTERSPEECH}}
  (\bibinfo{year}{2020}), \bibinfo{pages}{1808--1812}.
\newblock


\bibitem[\protect\citeauthoryear{Stappen, Schuller, Lefter, Cambria, and
  Kompatsiaris}{Stappen et~al\mbox{.}}{2020d}]%
        {stappen2020summary}
\bibfield{author}{\bibinfo{person}{Lukas Stappen}, \bibinfo{person}{Bj{\"o}rn
  Schuller}, \bibinfo{person}{Iulia Lefter}, \bibinfo{person}{Erik Cambria},
  {and} \bibinfo{person}{Ioannis Kompatsiaris}.}
  \bibinfo{year}{2020}\natexlab{d}.
\newblock \showarticletitle{Summary of MuSe 2020: Multimodal Sentiment
  Analysis, Emotion-target Engagement and Trustworthiness Detection in
  Real-life Media}. In \bibinfo{booktitle}{\emph{Proceedings of the 28th ACM
  International Conference on Multimedia}}. \bibinfo{pages}{4769--4770}.
\newblock


\bibitem[\protect\citeauthoryear{Stappen, Schumann, Sertolli, Baird, Weigel,
  Cambria, and Schuller}{Stappen et~al\mbox{.}}{2021c}]%
        {stappen2021toolbox}
\bibfield{author}{\bibinfo{person}{Lukas Stappen}, \bibinfo{person}{Lea
  Schumann}, \bibinfo{person}{Benjamin Sertolli}, \bibinfo{person}{Alice
  Baird}, \bibinfo{person}{Benjamin Weigel}, \bibinfo{person}{Erik Cambria},
  {and} \bibinfo{person}{Bj{\"o}rn~W Schuller}.}
  \bibinfo{year}{2021}\natexlab{c}.
\newblock \showarticletitle{MuSe-Toolbox: The Multimodal Sentiment Analysis
  Continuous Annotation Fusion and Discrete Class Transformation Toolbox}. In
  \bibinfo{booktitle}{\emph{Proceedings of the 2nd Multimodal Sentiment
  Analysis Challenge, co-located with the 29th ACM International Conference on
  Multimedia (ACMMM)}}. ACM, \bibinfo{address}{Changu, China}.
\newblock


\bibitem[\protect\citeauthoryear{Sun, Lian, Tao, Liu, and Niu}{Sun
  et~al\mbox{.}}{2020}]%
        {sun2020multi}
\bibfield{author}{\bibinfo{person}{Licai Sun}, \bibinfo{person}{Zheng Lian},
  \bibinfo{person}{Jianhua Tao}, \bibinfo{person}{Bin Liu}, {and}
  \bibinfo{person}{Mingyue Niu}.} \bibinfo{year}{2020}\natexlab{}.
\newblock \showarticletitle{Multi-modal Continuous Dimensional Emotion
  Recognition Using Recurrent Neural Network and Self-Attention Mechanism}. In
  \bibinfo{booktitle}{\emph{Proceedings of the 1st International on Multimodal
  Sentiment Analysis in Real-life Media Challenge and Workshop}}.
  \bibinfo{pages}{27--34}.
\newblock


\bibitem[\protect\citeauthoryear{Thelwall, Buckley, Paltoglou, Cai, and
  Kappas}{Thelwall et~al\mbox{.}}{2010}]%
        {thelwall2010sentiment}
\bibfield{author}{\bibinfo{person}{Mike Thelwall}, \bibinfo{person}{Kevan
  Buckley}, \bibinfo{person}{Georgios Paltoglou}, \bibinfo{person}{Di Cai},
  {and} \bibinfo{person}{Arvid Kappas}.} \bibinfo{year}{2010}\natexlab{}.
\newblock \showarticletitle{Sentiment strength detection in short informal
  text}.
\newblock \bibinfo{journal}{\emph{Journal of the American Society for
  Information Science and Technology}} \bibinfo{volume}{61},
  \bibinfo{number}{12} (\bibinfo{year}{2010}), \bibinfo{pages}{2544--2558}.
\newblock


\bibitem[\protect\citeauthoryear{Valstar, Schuller, Smith, Eyben, Jiang,
  Bilakhia, Schnieder, Cowie, and Pantic}{Valstar et~al\mbox{.}}{2013}]%
        {valstar2013avec}
\bibfield{author}{\bibinfo{person}{Michel Valstar}, \bibinfo{person}{Bj{\"o}rn
  Schuller}, \bibinfo{person}{Kirsty Smith}, \bibinfo{person}{Florian Eyben},
  \bibinfo{person}{Bihan Jiang}, \bibinfo{person}{Sanjay Bilakhia},
  \bibinfo{person}{Sebastian Schnieder}, \bibinfo{person}{Roddy Cowie}, {and}
  \bibinfo{person}{Maja Pantic}.} \bibinfo{year}{2013}\natexlab{}.
\newblock \showarticletitle{AVEC 2013: the continuous audio/visual emotion and
  depression recognition challenge}. In \bibinfo{booktitle}{\emph{Proceedings
  of the 3rd ACM International Workshop on Audio/Visual Emotion Challenge}}.
  ACM, \bibinfo{pages}{3--10}.
\newblock


\bibitem[\protect\citeauthoryear{W{\"o}llmer, Eyben, Reiter, Schuller, Cox,
  Douglas-Cowie, and Cowie}{W{\"o}llmer et~al\mbox{.}}{2008}]%
        {wollmer2008abandoning}
\bibfield{author}{\bibinfo{person}{Martin W{\"o}llmer},
  \bibinfo{person}{Florian Eyben}, \bibinfo{person}{Stephan Reiter},
  \bibinfo{person}{Bj{\"o}rn Schuller}, \bibinfo{person}{Cate Cox},
  \bibinfo{person}{Ellen Douglas-Cowie}, {and} \bibinfo{person}{Roddy Cowie}.}
  \bibinfo{year}{2008}\natexlab{}.
\newblock \showarticletitle{Abandoning emotion classes-towards continuous
  emotion recognition with modelling of long-range dependencies}. In
  \bibinfo{booktitle}{\emph{Proceedings of INTERSPEECH}}.
  \bibinfo{pages}{597--600}.
\newblock


\bibitem[\protect\citeauthoryear{Yang, Luo, Loy, and Tang}{Yang
  et~al\mbox{.}}{2015}]%
        {DBLP:journals/corr/YangLLT15b}
\bibfield{author}{\bibinfo{person}{Shuo Yang}, \bibinfo{person}{Ping Luo},
  \bibinfo{person}{Chen~Change Loy}, {and} \bibinfo{person}{Xiaoou Tang}.}
  \bibinfo{year}{2015}\natexlab{}.
\newblock \showarticletitle{{WIDER} {FACE:} {A} Face Detection Benchmark}.
\newblock \bibinfo{journal}{\emph{CoRR}}  \bibinfo{volume}{abs/1511.06523}
  (\bibinfo{year}{2015}).
\newblock
\showeprint[arxiv]{1511.06523}
\urldef\tempurl%
\url{http://arxiv.org/abs/1511.06523}
\showURL{%
\tempurl}


\bibitem[\protect\citeauthoryear{Zadeh, Morency, Liang, and Poria}{Zadeh
  et~al\mbox{.}}{2020}]%
        {challengehml2020grand}
\bibfield{editor}{\bibinfo{person}{Amir Zadeh}, \bibinfo{person}{Louis-Philippe
  Morency}, \bibinfo{person}{Paul~Pu Liang}, {and} \bibinfo{person}{Soujanya
  Poria}} (Eds.). \bibinfo{year}{2020}\natexlab{}.
\newblock \bibinfo{booktitle}{\emph{Second Grand-Challenge and Workshop on
  Multimodal Language (Challenge-HML)}}.
\newblock


\bibitem[\protect\citeauthoryear{Zhang, Zhang, Li, and Qiao}{Zhang
  et~al\mbox{.}}{2016}]%
        {zhang2016mtcnn}
\bibfield{author}{\bibinfo{person}{Kaipeng Zhang}, \bibinfo{person}{Zhanpeng
  Zhang}, \bibinfo{person}{Zhifeng Li}, {and} \bibinfo{person}{Yu Qiao}.}
  \bibinfo{year}{2016}\natexlab{}.
\newblock \showarticletitle{Joint Face Detection and Alignment Using Multitask
  Cascaded Convolutional Networks}.
\newblock \bibinfo{journal}{\emph{IEEE Signal Processing Letters}}
  \bibinfo{volume}{23} (\bibinfo{date}{04} \bibinfo{year}{2016}).
\newblock


\bibitem[\protect\citeauthoryear{Zhou and De~la Torre}{Zhou and De~la
  Torre}{2015}]%
        {zhou2015generalized}
\bibfield{author}{\bibinfo{person}{Feng Zhou} {and} \bibinfo{person}{Fernando
  De~la Torre}.} \bibinfo{year}{2015}\natexlab{}.
\newblock \showarticletitle{Generalized canonical time warping}.
\newblock \bibinfo{journal}{\emph{IEEE Transactions on Pattern Analysis and
  Machine Intelligence}} \bibinfo{volume}{38}, \bibinfo{number}{2}
  (\bibinfo{year}{2015}), \bibinfo{pages}{279--294}.
\newblock


\end{thebibliography}


\end{document}